\pdfoutput=1
\documentclass[11pt]{article}

\usepackage{acl}

\usepackage{times}
\usepackage{latexsym}

\usepackage{colortbl}

\usepackage[T1]{fontenc}
\usepackage[utf8]{inputenc}

\usepackage{microtype}

\usepackage{inconsolata}

\usepackage{enumitem}
\setlist{leftmargin=2.5mm}

\usepackage{xkeyval}

\usepackage{graphicx}

\usepackage{multirow}
\usepackage{xcolor} 
\usepackage{pifont}

\usepackage{lipsum} 

\newcommand{\benchie}{\textit{BenchIE}}
\newcommand{\brand}{\textit{B50}}
\newcommand{\bfl}{\textit{BenchIE}^{FL}}

\newcommand{\bflc}{\textit{BenchIE}^{FL}_{match}}

\newcommand{\wafl}{\textit{WebAssertions}^{FL}}
\newcommand{\wa}{\textit{WebAssertions}}

\newcommand{\sep}{~\textsc{--}~}

\newcommand{\cmark}{\ding{51}}%
\newcommand{\xmark}{\ding{55}}%
\definecolor{fecorrect}{HTML}{008000}
\definecolor{fewrong}{HTML}{e60000}

\newcommand{\correct}[1]{\cmark{} \textcolor{fecorrect}{#1}}
\newcommand{\wrong}[1]{\xmark{} \textcolor{fewrong}{#1}}
\newcommand{\fecomment}[1]{}

\title{\(\bfl\): A Manually Re-Annotated Fact-Based Open Information Extraction Benchmark}

\author{Fabrice Lamarche \and Philippe Langlais \\
        RALI, DIRO, Université de Montréal, Canada\\
         fabrice.lamarche@umontreal.ca\\
         felipe@iro.umontreal.ca}

\begin{document}
\maketitle
\begin{abstract}
Open Information Extraction (OIE) is a field of natural language processing that aims to present textual information in a format that allows it to be organized, analyzed and reflected upon. Numerous OIE systems are developed, claiming ever-increasing performance, marking the need for objective benchmarks. \benchie{} is the latest reference we know of. Despite being very well thought out, we noticed a number of issues we believe are limiting. Therefore, we propose \( \bfl \), a new OIE benchmark which fully enforces the principles of \benchie{} while containing fewer errors, omissions and shortcomings when candidate facts are matched towards reference ones. \( \bfl \) allows insightful conclusions to be drawn on the actual performance of OIE extractors.
\end{abstract}

\section{Introduction}

Open Information Extraction (OIE), the task of extracting organized tuples containing information expressed in a sentence \cite{yates-etal-2007-textrunner} has numerous downstream applications ranging from Question Answering \cite{10.1145/2623330.2623677} to Text Comprehension \cite{stanovsky-etal-2015-open}. 

Earlier works evaluated OIE extractors mainly by examining  their output and manually determining whether the extracted tuples were expressed in a given sentence. This method lacks the capacity to measure the recall of systems, which led to the creation of OIE benchmarks consisting of annotations of all possible tuples from a corpus and a matching function establishing the concordance between extracted and annotated facts. Our analysis of the most recent benchmark, \benchie{} \cite{gashteovski-etal-2022-benchie}, shows that although very well thought out, its results are noisy and prone to biases, making its conclusion less trustworthy.

% rule in Latex: never space by yourself: do not use \\ \\  or things like this. 

\paragraph{Contributions} 

Our main contribution in this work is the release of a new OIE benchmark, \(\bfl\), that we created by re-annotating \benchie, correcting frequent errors, inconsistencies and methodology limitations, resulting in more concise, precise and pertinent annotations. \(\bfl\) also benefits from a new matching function that is more flexible, and which captures more valid extractions, thus producing --- as we shall see --- a fairer ranking of evaluated systems. 

In doing so, we produce a number of useful resources\footnote{Resources, guidelines as well as evaluation scripts used for our experiments are available at \url{https://github.com/rali-udem/benchie_fl.git}.}, including new OIE guidelines, both for sentence annotation and tuple matching. We compare seven OIE extractors --- neural and non neural --- and show that believed state-of-the-art systems are not necessarily the best. We further demonstrate on three downstream tasks that the scores of extractors on \(\bfl\) exhibits a much stronger correlation with their performance on each task.
 
\paragraph{} This paper is organized as follows. In section~\ref{sec:related}, we review some existing OIE systems and benchmarks. We introduce issues with \benchie{} in section \ref{sec:issues}, leading us to define guidelines discussed in section~\ref{sec:guidelines}. In section~\ref{sec:benchie} we characterize \(\bfl\), a new version of \benchie{} we annotated by following our guidelines. Experiments are reported in section~\ref{sec:exp} and we conclude in section~\ref{sec:conclusion}.

% =====
% =====
\section{Related Work}
\label{sec:related}

\subsection{OIE Systems}
\label{sec:sys}

Open information extraction systems vary in their approaches to the task. Earlier systems are mostly rule-based. These systems, like \textit{ReVerb} \cite{fader-etal-2011-identifying}, \textit{ClausIE} \cite{10.1145/2488388.2488420} or \textit{MinIE} \cite{gashteovski-etal-2017-minie} make use of parts-of-speech tags, syntactical analysis and other grammatical characteristics to derive simple extraction rules. Newer systems are almost exclusively using neural approaches, and belonging to one of two distinct categories, either sequence-to-sequence or sequence tagging techniques. Sequence-to-sequence models, like \textit{ImojIE} \cite{kolluru-etal-2020-imojie} and \textit{M2OIE} \cite{ro-etal-2020-multi} output the original sentence one word at a time, as well as markers that delimit the arguments and relations, making use of copying and attention mechanisms to output the original sentence words. Neural tagging models like \textit{OpenIE6} \cite{kolluru-etal-2020-openie6} and \textit{CompactIE} \cite{fatahi-bayat-etal-2022-compactie} use different approaches but mainly focus on identifying relations initially, and then tagging other words as arguments related to identified relations. Models from both categories are typically trained on bootstrapped data from previously released extractors  selected with confidence scores and heuristics.

Further information on those extractors and implementation details are provided in Appendix~\ref{app:implementation}.

\subsection{OIE Benchmarks}

The first complete OIE benchmark, \textit{OIE2016} \cite{stanovsky-dagan-2016-creating} was created by automatically transforming two Question Answer Driven Semantic Role Labeling (QA-SRL) datasets \cite{he-etal-2015-question} into a benchmark comprising 3200 sentences (from the Wall Street Journal and Wikipedia) and 10359 extractions. The matching function used by \textit{OIE2016} matches an extraction and a reference tuple if the grammatical head of both their arguments and their relation are the same.

\newcite{lechelle-etal-2019-wire57} identifyied limitations both in the conversion involved in \textit{OIE2016} and its matching function (they demonstrate that it is straightforward to attack this benchmark with a dummy extractor). They released \textit{WiRE57}, a benchmark  made up of 57 expertly annotated sentences from Wikipedia and Reuters into 343 tuples, and use \textbf{token-level} scoring, meaning that for each annotated tuple, the precision is measured as the proportion of extracted words present in the annotations, and the recall, the proportion of annotated words present in the extractions. \textit{WiRE57} is purposely criticized for its small size. 
 
 The authors of \textit{CaRB} \cite{bhardwaj-etal-2019-carb} proposed crowd-sourcing as a solution to the high cost of manually annotating sentences. They release their reference of 2714 tuples from a subset of 1200 sentences from \textit{OIE2016} annotated thanks to Amazon Mechanical Turk. They also introduce slight modifications to \textit{WiRE57}’s scoring function.
 
 \fecomment{ eliminating the penalties for long extractions and those  that combine information from multiple annotated tuples.}

\newcite{gashteovski-etal-2022-benchie} criticize \textit{CaRB} mainly in the way it scores system extractions; showing that token-level scoring allows for incorrect extractions to be scored highly both in precision and recall. To counter this, they propose to use a conservative \textbf{exact matching} function, meaning that  only extractions that are identical to an annotated tuple will count. This notion of exact match works because of the \textit{fact synset} principle they introduce~: instead of annotating only one formulation of a given fact, they aim to list all \textit{possible} formulations of the fact in a single \textit{synset} or \textit{cluster}. Thus, if an extraction matches any of the formulations of a \textit{synset}, is it said to match that \textit{cluster}. They manually annotate 300 sentences of the original \textit{OIE2016} dataset, resulting in 1354 \textit{clusters}. 

% -------- fig 1 ----------------
\begin{table*}[htbp]
\center
\resizebox{\textwidth}{!}{%
\begin{tabular}{|ll|}
\hline
\multicolumn{1}{|c|}{\textbf{Error type}} &
  \multicolumn{1}{c|}{{\color[HTML]{000000} \textbf{Annotation}}} \\ \hline \hline
  
\multicolumn{2}{|c|}{\textit{\begin{tabular}[c]{@{}c@{}}For example , when two such hydrophobic particles come very close , the clathrate-like baskets surrounding them merge\end{tabular}}} \\ \hline
\multicolumn{1}{|l|}{\textbf{Missing Fact}} &
  \multicolumn{1}{l|}{\begin{tabular}[c]{@{}c@{}}when two such hydrophobic particles\sep come very close\sep{[}the{]} clathrate-like baskets surrounding them merge\end{tabular}} \\ \hline
\multicolumn{1}{|l|}{\textbf{False Fact}} &
  two {[}such{]} hydrophobic particles \sep come \sep {[}very{]} close \\ \hline \hline

  \multicolumn{2}{|c|}{\textit{\begin{tabular}[c]{@{}c@{}}He and his friends were said to have made bombs for fun on the outskirts of Murray , Utah .\end{tabular}}} \\ \hline
\multicolumn{1}{|l|}{\textbf{Irrelevant fact}} &
  his friends\sep were said to have made bombs on \sep {[}the{]} outskirts {[}of Murray , Utah{]} \\ \hline \hline
  
\multicolumn{2}{|c|}{\textit{\begin{tabular}[c]{@{}c@{}}They held the first Triangle workshop for thirty painters from the US , the UK and Canada at Pine Plains , New York .\end{tabular}}} \\ \hline
\multicolumn{1}{|l|}{\begin{tabular}{l}\textbf{Double annotation}\\ \end{tabular}} &
  \begin{tabular}[c]{@{}l@{}} 
  They \sep held {[}the{]} {[}first{]} Triangle workshop at \sep {[}Pine Plains{]} New York\\ \\  
  They \sep held {[}the{]} {[}first{]} Triangle workshop at \sep New York
  \end{tabular} \\ \hline
\multicolumn{1}{|l|}{\begin{tabular}{l}\textbf{Double meaning}\\ \end{tabular}} &
  \begin{tabular}[c]{@{}l@{}}
  They  \sep  held {[}the{]} {[}first{]} Triangle workshop  \sep  for {[}thirty{]} painters from {[}the{]} US\\
  They  \sep  held {[}the{]} {[}first{]} Triangle workshop  \sep  for {[}thirty{]} painters from Canada\end{tabular} \\ \hline
\end{tabular}%
}

\caption{Illustration of error types found in a sample of \benchie. Square brackets indicate words that are optional when matching facts.}
\label{table:ErrEx}

\end{table*}
% ----------------------

These  benchmarks yield different conclusions regarding the best performing extractors. This is because their annotation principles, text corpora and matching functions are all different. Recently, \newcite{Pei:2023} made recommendations for assisting in deciding the best (neural) extractor for a  given downstream task, and which benchmarks' characteristics better correlate with it. While we agree that ultimately, we should test extractors on specific tasks (which we also revisit in Section~\ref{sec:abqa}), there is a need for sound references that will help appreciate limitations of current extraction technology, hopefully leading to better extractors. In their study, \newcite{Pei:2023} rejected \benchie{} in comparing benchmarks,\footnote{On the grounding that it is a fact-centered  benchmark, which we see as a positive characteristic.} while our inspections make us believe it is the most well-though benchmark.\footnote{ Our review of existing benchmarks is not exhaustive. In particular, \textit{ReOIE2016} \cite{zhan2019span} and \textit{LSOIE} \cite{solawetz-larson-2021-lsoie} have been proposed as updated versions of \textit{OIE2016}, but they both use the same matching function, which has been largely criticized.}

% =====
% =====
\section{Issues with \benchie}
\label{sec:issues}

%\subsection*{To place at the right place}
%
%\begin{itemize}
%
%\item \felipe{Regarding inference, \newcite{lechelle-etal-2019-wire57} argued that some facts not implicit in a sentence might be inferred. While arguable, we do believe this is a good practice. See Appendix \ref{sec:appendixannoG} for guidelines regarding annotation.}{to integrate carefully}
% 
% \end{itemize}

By inspecting a (random) sample of 50 sentences of \textit{BenchIE} - hereafter referred to as \brand{} - we noticed a number of issues we think worth being taken care of for more meaningful comparisons of extractors.  

\subsection{Annotation problems}
\label{sec:annProb}
 
We identified five error types that are illustrated in Table~\ref{table:ErrEx}.

  \paragraph{\textbf{Missing fact}} A fact is missing from the reference. This may be due to the lack of inference in \textit{BenchIE} (see section \ref{sec:guidelinesAnno}), or from oversight or omission. The example shows the main information piece from the sentences and no formulation of that fact is present in the annotated clusters.
  
  %This fact is not a case of inference, it is simply missing from the annotations. 

  \paragraph{\textbf{False fact}} A fact in the reference is false, or not necessarily implied by the sentence. The example shows a fact that is false because it lacks the mention that \textit{hydrophobic particles} \underline{can} \textit{come very close} (when \textit{baskets surrounding them merge}), instead conveying the fact that they \underline{do}.

  \paragraph{\textbf{Irrelevant fact}} A fact in the reference lacks relevance, because of missing context or other information. In the example, without the optional \textit{Murray, Utah}, the fact is not relevant because \textit{bombs} being made on \textit{outskirts} is not relevant information.

  \paragraph{\textbf{Double annotation}} Two clusters have at least one formulation that expresses the same information. The example shows two clusters being exactly the same without the optional words. The optional words in the first cluster should be \textit{New York} and not \textit{Pine Plains}.

   \paragraph{\textbf{Double meaning}} A single cluster has at least two formulations that express different information. The example shows two formulations of the same cluster conveying very distinct information.

\paragraph{} Table~\ref{table:ErrFreq} shows the number of sentences in \brand{} for which a cluster or a formulation shows a given error type. We  observe that more than half the sentences have one or more missing facts. In many cases, this is because of inference (see section \ref{sec:guidelinesAnno}), but some facts have simply been omitted. Both irrelevant or false facts are present in about a third of the sentences. This high frequency of issues in \benchie{} shows the need for a more thorough annotation of the original sentences. These new annotations need to be motivated and conducted by solid guidelines that are typically lacking in the OIE task.

\subsection{Matching problem} 

Alongside the errors found in \benchie's annotations, we annotated facts output by \textit{ReVerb}, \textit{IMojIE}, \textit{OpenIE6}, and \textit{CompactIE} and found  many cases were a system made an extraction that in our eye was valid and should match an annotated fact that was not matched by the exact match used in the \benchie{} benchmark.  This is because while \newcite{gashteovski-etal-2022-benchie}  argue that they listed all possible valid formulations of a given fact in each cluster, we find that this is not the case (and we argue that this is in practice very hard to do).

 In the 50 sentences  of \brand, we found 26  (52\%) sentences with at least one fact from one of the extractors that was not matched when it should have. Because of this, we develop a new matching function aiming to capture more matches between extracted and annotated tuples.

\begin{table}[htbp]
\centering
\begin{tabular}{| 
>{\columncolor[HTML]{FFFFFF}}c 
>{\columncolor[HTML]{FFFFFF}}r 
>{\columncolor[HTML]{FFFFFF}}r |}
\hline
\textbf{Error type}      & \textbf{Count}  & \textbf{\%}\\ \hline \hline
Missing Fact             & 26                    & 52          \\  
False fact                  & 15                    & 30       \\  
Irrelevant fact             & 32                  & 64        \\ 
Double annotation      & 28                  & 56            \\ 
Double meaning         & 15                  & 30            \\ \hline
\end{tabular}
\caption{Count of sentences with a given error type in the 50 sentences of \brand{} in \benchie.}
\label{table:ErrFreq}
\end{table}

% ----- Fig 2 ------
\begin{table*}[htbp]
\resizebox{\textwidth}{!}{%
\begin{tabular}{|l|l|l|}
\hline
\multicolumn{1}{|c|}{\textbf{Principle}} &
  \multicolumn{1}{c|}{\textbf{Sentence}} &
  \multicolumn{1}{c|}{\textbf{Annotations}} \\ \hline \hline
Informativity &
  \textit{Alex lived in Paris and is now living in Cologne.} &
  \begin{tabular}[c]{@{}l@{}}
  \correct{Alex \sep is now living in \sep Cologne}\\ 
  \wrong{Alex \sep is now \sep living}
  \end{tabular} \\ \hline \hline
Minimality &
  \textit{The group was created in 2020 by three people.} &
  \begin{tabular}[c]{@{}l@{}}
  \correct{The group \sep was \sep created}\\ 
  \correct{The group \sep was created in \sep 2020}\\ 
  \correct{The group  \sep was created by \sep three people}\\ 
  \wrong{The group \sep was created \sep in 2020 by three people}
  \end{tabular} \\ \hline \hline
Exhaustivity &
  \textit{\begin{tabular}[c]{@{}l@{}}He has written several newspaper\\ and magazine opinion pieces.\end{tabular}} &
  \begin{tabular}[c]{@{}l@{}}
  \correct{He \sep has written \sep {[}several{]} pieces}\\
  \correct{He \sep has written \sep {[}several{]} opinion pieces}\\ 
  \correct{He \sep has written \sep {[}several{]} newspaper pieces}\\ 
  \correct{He \sep has written \sep {[}several{]} magazine pieces}\end{tabular} \\ \hline \hline
Relation completeness &
  \textit{Tokyo’s population is over 13 millions.} &
  \begin{tabular}[c]{@{}l@{}}
  \correct{Tokyo’s population \sep is over \sep 13 millions}\\ \
  \wrong{Tokyo’s population \sep \textbf{is} \sep over 13 millions}
  \end{tabular} \\ \hline \hline
Inference &
  \textit{\begin{tabular}[c]{@{}l@{}}
  ‘My classical way’ was released in 2010 \\ 
  on Marc’s own label, Frazzy Frog Music.
  \end{tabular}} &
  \correct{Marc’s {[}own{]} label \sep is \sep Frazzy Frog Music} \\ \hline
\end{tabular}%
}
\caption{Illustration of annotation principles. Facts in \textcolor{fecorrect}{green} (preceded by a check mark) should be included in the annotation, while facts in \textcolor{fewrong}{red} (preceded by an cross mark) should not.} 
\label{table:exguidelinesanno}
\end{table*}
% ------------------------
% =====
% =====
\section{Guidelines}
\label{sec:guidelines}

\subsection{Annotation Guidelines}
\label{sec:guidelinesAnno}

Few principles are universally accepted by OIE benchmarks and systems authors. Here, we try to list crucial principles that make most sense for OIE output to be useful for downstream tasks, and aim for those to guide annotation of our and future references. Our full annotation guidelines, along with examples illustrating the principles can be found in Appendix~\ref{sec:appendixannoG}.  Examples for the following principles are shown in Table~\ref{table:exguidelinesanno}.
 
\paragraph{\textbf{Informativity}} Tuples should be informative and not contain generalities. This general principle, also present in \benchie, means that as long as a tuple contains an information expressed in the sentence and that it respects all other principles, it should be included in the annotations. In the negative example, the second argument \textit{living} is not informative, thus that fact should not be annotated.
 
\paragraph{\textbf{Minimality}}  Each tuple should be minimal, meaning that it can not be separated into multiple distinct tuples. The faulty annotation in line 2 of Table~\ref{table:exguidelinesanno} combines three facts, and should therefore not be annotated. No mention of minimality is made in \benchie, and their guidelines only suggest making non-necessary words optional, which in practice allows for overly long and imprecise annotations.
  
\paragraph{\textbf{Exhaustivity}} The set of tuples for a given sentence should cover all pieces of information expressed in the sentence. At the cluster level, all of the possible formulations for which any of the arguments or the relation is different should be listed. In the example, all the types of \textit{writing} that \textit{He} has done should be annotated in separate clusters. \textit{BenchIE}'s guidelines try to handle exhaustivity by listing all verb-mediated facts, which does not capture all information.

\paragraph{\textbf{Relation Completness}} Relations are responsible for the information; as such they should be complete, meaning that the information in the arguments do not change the core meaning of the relation. In the example, the negative fact should not be annotated because its relation, \textit{is}, is not complete, its meaning is modified by the word \textit{over} in the second argument. In contrast, \benchie 's guideline specifically encourages annotators to place words that are not verbs in or out of relations without regard for relation integrity.

\paragraph{\textbf{Inference}} We define inferred tuples as facts that are implied by the sentence (true if the sentence is true), but for which the relation linking the arguments is not present in the text. Inference should be carried out because it is useful in downstream tasks such as QA or knowledge base (KB) population \cite{gashteovski-etal-2020-aligning}. However, limits should be set in regards to the information inferred. These nuances are explained in detail in the complete guidelines but simply put, tuples that can be inferred without needing complex reasoning or external knowledge should be annotated. In the last example of Table \ref{table:exguidelinesanno}, the tuple is included because it is clearly implied by the sentence. Since \textit{BenchIE} only includes facts mediated by words present in the sentence, they annotate almost no inferred facts.

% ------ Fig 3 -----------
\begin{table*}[htbp]
\resizebox{\textwidth}{!}{%
\begin{tabular}{|l|l|l|}
\hline
\multicolumn{1}{|c|}{\textbf{Principle}} &
  \multicolumn{1}{c|}{\textbf{Sentence}} &
  \multicolumn{1}{c|}{\textbf{Annotations}} \\ \hline \hline
\multirow{2}{*}{Relation specificity} &
  \multirow{2}{*}{\textit{The party was thrown out of the government.}} &
  \begin{tabular}[c]{@{}l@{}}The party \sep \textbf{was} \sep thrown out of {[}the{]} government\\ The party \sep \textbf{was thrown out of} \sep {[}the{]} government\end{tabular} \\ \cline{3-3} 
 &
   &
   \wrong{The party \sep \textbf{was thrown} \sep out of {[}the{]} government} \\ \hline
\multirow{2}{*}{Word choice} &
  \multirow{2}{*}{\textit{He is the older brother of Alex.}} &
   He \sep is \sep \textbf{{[}an{]}} older brother \\ \cline{3-3} 
 &
   &
  \correct{He \sep is \sep \textbf{the} older brother} \\ \hline
\multirow{2}{*}{Level of detail} &
  \multirow{2}{*}{\textit{Alex broadcasts a web series Music on a website.}} &
  \begin{tabular}[c]{@{}l@{}}Alex \sep broadcasts, {[}a{]} web series\\ Alex \sep broadcasts, Music\\ Alex \sep broadcasts Music on, a website\end{tabular} \\ \cline{3-3} 
 &
   &
  \begin{tabular}[c]{@{}l@{}} 
  \correct{Alex \sep broadcasts \sep Music on a website}\\  
  \wrong{Alex \sep broadcasts \sep a web series Music on a website}\end{tabular} \\ \hline
\end{tabular}%
}
\caption{Examples of matching principles. Annotated facts are in black, facts in \textcolor{fecorrect}{green} (preceded by a check mark) should be matched to the annotations, while facts in \textcolor{fewrong}{red} (preceded by a cross mark) should not.} 
\label{table:exguidelinesmatch}
\end{table*}

%(marked by \cmark)
%(marked by  \xmark)
% -----------------------

\subsection{Matching Guidelines}

 To guide which extracted tuples should match with which annotated clusters, we need matching guidelines. Exactly identical extraction and annotation should obviously match and these cases are the only matches scored in \textit{BenchIE}. However, we believe other nuances exist and the following principles aim to illustrate these nuances. Examples are shown in Table \ref{table:exguidelinesmatch}. Our full matching guidelines can be found in Appendix~\ref{sec:appendixmatchG}.

  \paragraph{\textbf{Relation specificity}} The extracted and annotated relations should be as specific, meaning that prepositions and linking words should not be arbitrarily placed in the arguments or in the relation. In the example, the relation \textit{was thrown} is not as precise and complete as the annotated relations and its meaning is changed by the words \textit{out of} in the second argument, thus the extraction should not be matched.

  \paragraph{\textbf{Word choice}}  Certain  extractions  may contain syntax errors, misplaced words or other word choices that are different to those of the annotations. These are not inherently bad but if they affect the sense of the extraction, then the extraction should not match. The example shows the case were even if the extraction and the annotation do not use the same determiner (\textit{an} and \textit{the}), they both convey the same meaning, i.e. that \textit{He} is older than his brother.

  \paragraph{\textbf{Level of detail}} Many system extractions carry information from more than one annotated tuple. We want to match these extractions only if the extraction combines information from no more than two clusters, otherwise, we consider it too noisy for a  downstream task. The example shows extractions that combine information from two and from three clusters respectively. The negative example is noisier since its third argument is too long and lacks  preciseness.

\begin{table}[htbp]
\resizebox{\columnwidth}{!}{%
\centering
\begin{tabular}{|l|l|l|}
\hline
\multicolumn{1}{|c|}{} & {\color[HTML]{000000} \textbf{\(\bfl\)}} & \textbf{\textit{BenchIE}} \\ \hline \hline
Total clusters          & 1798  & 1354  \\  
Avg cluster/sentence    & 6  & 4.5  \\ 
Avg formulation/cluster & 3  & 6  \\  
Avg formulation length  & 10.6 & 12.5 \\  
%Med formulation length  & 9     & 12    \\ \hline
Avg relation length     & 3.9  & 4.0  \\ \hline
%Med relation length     & 3     & 3     \\ \hline
\end{tabular}%
}
\caption{Annotation statistics of \benchie{} and our re-annotated version: \(\bfl\).} 
\label{table:caracteristiquesanno}
\end{table} 

% =====
% =====
\section{\(\bfl\)}
\label{sec:benchie}

Following our new annotation guidelines, the first author of this paper (NLP scientist with 2 years of expertise) annotated \benchie{}’s original sentences resulting in \(\bfl\), a re-annotated corpus of 300 sentences. This  annotation effort was conducted using the \textit{AnnIE} annotation platform \cite{friedrich-etal-2022-annie}. 
 
Different statistics regarding both annotation sets are reported in Table~\ref{table:caracteristiquesanno}. First, we annotate more facts in total, that is, a higher average number of clusters per sentence. This is because we include inferred information and follow the minimality principle (meaning that we divide the information as much as possible). Second, we annotate (far) fewer formulations per cluster, both because relationship specificity is of great importance in our guidelines, and because we don't rely on an exact match function (see Section~\ref{sec:matching}), so there is no need for \(\bfl\) to list all possible formulations of the same fact.  Third, we note that on average,  our annotations are shorter; again due to the  minimality principle. On the other hand, the mean  lengths of the relations are more or less equivalent, due to our desire to preserve the specificity of the relations. Appendix \ref{sec:appendixanno} shows an example highlighting differences in both annotations of the same sentence.

%Overall, these figures suggest that annotating a resource according to our guidelines requires less work.
% felipe: pas sur qu'on peut deduire cela de ces chiffres, 
% dans le doute enlevons
% We can therefore hypothesize that our annotations are more concise, better summarizing the information while retaining the specificity and informativeness of the relations, which is one of the main objectives of our re-annotation.

\subsection{Manual evaluation}
\label{sec:manEval}

%Since all of our annotations are carried out by a single annotator, we set out to validate them using two different annotators. We present them with the first 50 sentences and their annotations from both sets of annotations (25 each), and then ask them to (blindly) annotate :  \textbf{exhaustivity} by indicating if a set of annotations for a given sentence fail to covers all the facts expressed, \textbf{minimality} by indicating clusters that  can be separated into several (smaller) clusters, and \textbf{relation completeness} by marking relations that are  modified by their arguments or that do not hold on their own. The results of the validation experiment are presented in Table \ref{table:valexp} where each cell represents the count of error-full sentences for each criteria by each annotator for both sets of annotations.

Since all of our annotations are carried out by a single annotator, we set out to validate them using two other annotators (NLP scientists with over 20 years of expertise, $a_2$ not being a co-author).  We present them with sentences of \brand{} and their annotations from both sets of annotations (25 each)\footnote{See Appendix \ref{app:infereds} for details of slight modifications we made to our annotations for this experiment.}, and then ask them to (blindly) annotate : \textbf{exhaustivity} by indicating if a set of annotations for a given sentence fail to covers all the facts expressed, \textbf{minimality} by indicating clusters that  can be separated into several (smaller) ones, and \textbf{relation completeness} by marking relations that are  modified by their arguments or that do not hold on their own.

\begin{table}[hbtp]
\resizebox{\columnwidth}{!}{%
\begin{tabular}{l|ll|ll|l}
\cline{2-5}
                                   & \multicolumn{2}{l|}{\textbf{\(\bfl\)}} & \multicolumn{2}{l|}{\textbf{BenchIE}} &                                \\ \cline{2-6} 
                                   & \multicolumn{1}{r|}{$a_1$}  & \multicolumn{1}{r|}{$a_2$} & \multicolumn{1}{r|}{$a_1$} & \multicolumn{1}{r|}{$a_2$} & \multicolumn{1}{l|}{$\kappa$} \\ \hline
\multicolumn{1}{|l|}{Exhaustivity} & \multicolumn{1}{p{0.8cm}|}{\hfill 5}   & \multicolumn{1}{r|}{9}   & \multicolumn{1}{r|}{7}  & \multicolumn{1}{r|}{14} & \multicolumn{1}{l|}{0.62}      \\ \hline
\multicolumn{1}{|l|}{Minimality}   & \multicolumn{1}{r|}{3}   & \multicolumn{1}{r|}{16}  & \multicolumn{1}{r|}{5}  & \multicolumn{1}{r|}{9}  & \multicolumn{1}{l|}{0.68}      \\ \hline
\multicolumn{1}{|l|}{Incomplete relations}    & \multicolumn{1}{r|}{4}   & \multicolumn{1}{r|}{14}  & \multicolumn{1}{r|}{9}  & \multicolumn{1}{r|}{18} & \multicolumn{1}{l|}{0.68}      \\ \hline
\end{tabular}%
}
\caption{Validation  statistics. Non-minimal sentences are those with at least one non-minimal annotation, and  incomplete sentences are sentences with at least one cluster whose relation is incomplete. $a_1$ and $a_2$ represent the first and second annotators respectively.} 
\label{table:valexp}
\end{table}

Results are presented in Table \ref{table:valexp} where each cell represents the count of error-full sentences for each criteria by each annotator for both sets of annotations. We observe that for all criterion, the annotators found our annotations to follow the guidelines much more closely. Both annotators find more than double the errors in \textit{BenchIE} compared to our annotation for almost all error types. Table \ref{table:valexp} also shows Cohen's kappa scores for the agreement between the two annotators, which indicate a moderate agreement. We found validating annotations requires annotators to fully embrace guidelines, a too demanding task. We believe that OIE benchmarks are better evaluated by how strongly their results resemble performance of systems on downstream tasks, as explored in section \ref{sec:abqa}. In Appendix \ref{app:proxys}, we also devise (objective) tests that highlight annotation issues and show that \textit{BenchIE}'s annotations contain a lot more potential errors than ours.

% fabrice: est-ce qu'on peut montrer une seule ligne pour sauver de l'espace?
% --> j'en ai enlevée une, à confirmer
\begin{table*}[htbp]
\resizebox{\textwidth}{!}{%
\begin{tabular}{|l|l|c|}
\hline
\multicolumn{1}{|c|}{\textbf{System extraction}}          & \multicolumn{1}{c|}{\textbf{Annotation cluster}} & \textbf{Match} \\ \hline \hline
He \sep served as first Prime Minister of \sep Australia        & He \sep served as \sep {[}the{]}{[}a{]} Prime Minister & 0              \\ \hline
He \sep became \sep founding justice of High Court of Australia & He \sep became \sep {[}a{]} founding justice           & 1     \\ \hline
%...                                                       & ...                                              & ...            \\ \hline
\end{tabular}%
}
\caption{Examples of matching annotations. Index of clusters are not relevant here.} 
\label{table:exmatch}
\end{table*}

\subsection{Matching function}
\label{sec:matching}
As a reminder, previous OIE benchmarks (\textit{WiRE57}, \textit{CaRB} and others) used \textbf{token-level} matching, where a candidate extraction would be scored in precision and recall by the number of words overlapping with the best-matching annotation. \textit{BenchIE}'s authors introduced the \textbf{exact} matching, only possible thanks to \textit{synsets}, where a candidate extraction only matches an annotation if they are identical. We set out to develop a new matching function that captures more matches than the exact matching used in \benchie. In order to evaluate this matching function, we annotate the extractions produced by the seven systems introduced in Section~\ref{sec:sys} for the 50 sentences of  \brand, and indicate for each extraction the index of the cluster it should match to according to our matching guidelines; an extraction that should not match any cluster is marked by 0.  Examples of such annotations are provided in Table~\ref{table:exmatch}. The resulting resource, named  \(\bflc\) contains 9400 extraction-annotation pairs, 96.8\% of which have no associated cluster.

%We did similarly with the sentences of \textit{Wire57} \cite{lechelle-etal-2019-wire57} in order to get an out-domain set of matches to eventually train a matching function on, leading to \(\wflc\), a collection of 10973 extraction-annotation pairs (97.5\% non-match). Note that this required to first transform the annotations of  \textit{Wire57} into the  \benchie{} format (since it is not annotated with clusters of facts) and insuring that the resulting annotations satisfy our annotation guidelines. Since \textit{Wire57}  is a benchmark which gained in popularity lately, we also share \(\wfl\), the resulting resource.

%While  annotating  \(\wflc\), we gathered three simple types of heuristics that we found useful for accommodating extractions that merge information from several clusters.

We then gathered three simple types of heuristics that we found capture more of those annotated matches that are not exact matches :

 \paragraph{\textbf{Alternative formulations (AF)}}  In two specific (yet frequent) situations, we do credit an extraction which does not match a reference cluster because it regroups information from two clusters: when its relation is reducible\footnote{We say $r$ reduces to $r'$, if removing optional words from r leads to r'.} to \texttt{is}, and when one of its argument contains a coordinate conjunction \texttt{and}. Implementation details are provided in Appendix~\ref{app:matching}, but as an illustration, the tuple (He \sep is \sep Canadian \texttt{and} a musician) might yield to two alternatives (He \sep is \sep Canadian) and (He \sep is \sep a musician), that will more likely (exact) match the reference.

% We identify two cases where combining information from two distinct clusters help matching. The \textbf{is} case correspond to a situation where we have  \( (X , is,  Y) \) and  in a different cluster \( (X , rel,  Z) \). We can then add the formulation \( (Y , rel,  Z) \) to the second cluster. Similarly, the case \textbf{and} prescribes to add  formulations \( (X , rel,  Z) \) and \( (Y , rel,  Z) \) to a cluster which contains  \( (X\, and\, Y , rel,  Z) \). Those two rules  allow to capture matches that respect all matching principles but that do not have identical formulations in the reference. 

  \paragraph{\textbf{Level of detail Matching (LoD)}}  We match an extraction which  linearization\footnote{By linearization of a tuple, we mean the string obtained by concatenating in that order arg1, relation and arg2.}  is verbatim the one of a reference formulation and which one of its argument and its relation are also present in another cluster. A typical example is illustrated in Table~\ref{table:exguidelinesmatch} where the candidate tuple  (Alex \sep broadcasts \sep Music on a website) is matched because it has the same linearization as  the reference tuple (Alex \sep broadcasts Music on \sep a website) and \textit{Alex}, and \textit{broadcast} are found in the second cluster: (Alex \sep broadcast \sep Music) in the corresponding slots.

  % fabrice : Est-ce que c'est une bonne idée d'illustrer cette variante du match en utilisant l'exemple de notre guide de matching?
  % I guess...

  % felipe: why not !?

  %We match an extraction which  linearization\footnote{By linearization of a tuple, we mean the string obtained by concatenating in that order arg1, relation and arg2.}  is verbatim the one of a reference formulation and which relation is also present in another cluster. A typical example is illustrated in Table~\ref{table:exguidelinesmatch} where the candidate tuple  (Alex, broadcasts, Music on a website) is matched because it has the same linearization as  the reference tuple (Alex, broadcasts Music on, a website) and broadcast is a relation found in the second cluster: (Alex, broadcast, Music).
    
\paragraph{\textbf{Punctuation (Punc)}} We carry out matching removing all punctuation characters and lowercasing strings. This is because we have not listed all possible combinations of capitalization and punctuation in our annotations, and consider we should not. 
   
\paragraph{} Both AF and LoD heuristics help identify matches that occur because of the \textit{Level of Detail} principle from our guidelines, while the Punc heuristic applies to the \textit{Word Choice} principle. The F1 score of matching function variants on \(\bflc\) is reported in Table~\ref{table:perfcorr}. We observe that the exact match function is outperformed by each addition of the heuristics we described, and that using them all leads to the best performance. Therefore we selected (EM + AF + LoD + Punc) as the default matching function in the evaluation toolkit accompanying \(\bfl\). We further verify in Appendix~\ref{app:comparison_sf}, that the ranking of systems thanks to  our scoring function correlates better than other scoring functions to the ranking obtained by evaluating systems based on the (manual) matching annotations we conducted.\footnote{We also attempted training a matching function on \(\bflc\), but found disapointing result with low generalization on unseen matching pairs.}

\begin{table}[htbp]
\begin{center}
%\resizebox{\columnwidth}{!}{%
\begin{tabular}{|l|l|l|}
\hline
\multicolumn{1}{|c|}{\textbf{Matching function}}                               & \multicolumn{1}{c|}{{\color[HTML]{000000} \textbf{F1}}} \\ \hline \hline
Exact Match (EM)                                                                & 0.83 \\ \hline \hline
EM + AF 										& 0.87 \\  
EM + LoD 									& 0.87 \\  
EM + AF + LoD                                                                    & 0.96 \\  
EM + AF + LoD + Punc 							& \textbf{0.97}     \\ \hline
\end{tabular}%}
\end{center}

\caption{Matching F1 of matching function variants on \(\bflc\) .}
\label{table:perfcorr}
\end{table} 

%We also train classifiers on  \(\wflc\), both multi-layer perceptrons and random forests using LightGBM on different sets of features, ranging from matching word counts and similarity metrics to specific word usage and token-level matching scores, with a total of over 50 features created. While the perceptron performs very well on the data set it has been trained on, it does not generalize  much better than the random forest classifier on  \(\bflc\), and both classifiers significantly underperform the best  heuristic-based matching functions. 

% felipe: we could report a table with precision and recall in annexe and simply point to it here.

% ============
% ============
\section{Experiments}
\label{sec:exp}

We use the default configuration for  the 7 extractors mentioned in Section~\ref{sec:sys} without attempting to optimize them for specific benchmarks. \newcite{Pei:2023} observed that optimizing such systems do not lead to significant performance differences.

\begin{figure*}[htbp]
\center
\includegraphics[scale=0.315]{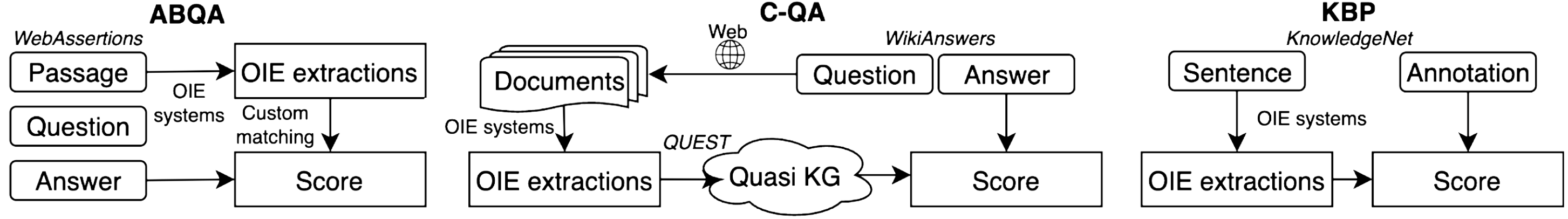}
\caption{Downstream tasks flowchart.} 
\label{fig:tasks_diagram}
\end{figure*}

\subsection{Downstream tasks}
\label{sec:abqa}

Ultimately, OIE extractions, and thus systems, are only useful in downstream applications. This means that a useful benchmark’s rankings should follow closely the rankings obtained by evaluating OIE extractions usefulness in these downstream tasks. To test whether this is the case for our reference, we study three tasks that directly use OIE triples: Assertion-Based Question Answering \cite{yan2018assertionbased}, Complex Questions Answering with quasi Knowledge Graphs \cite{Lu_2019} and Knowledge Base Population \cite{mesquita-etal-2019-knowledgenet}. See Figure \ref{fig:tasks_diagram} for a visual description of the tasks, Table~\ref{table:ScoresDT}\footnote{\textit{IMojIE} is exluded from C-QA and KBP because of compute constraints.} for systems scores and Annex \ref{app:pearson} for examples and details on each task's setup.

\begin{table}[htbp]
\centering
\begin{tabular}{|llll|}
\hline
\textbf{System}        & \textbf{ABQA} 	& \textbf{C-QA} & \textbf{KBP}\\ \hline \hline
\textit{ReVerb}         & 0.230      		&    0.092 &    0.149\\
\textit{ClausIE}         & 0.180      		&    0.089 &    0.026\\
\textit{MinIE}         & 0.270      		&    0.095  &    0.396\\ 
\textit{IMojIE}         & 0.170      		&    - &    - \\
\textit{OpenIE6}         & 0.170      		&    0.087 &    0.064\\
\textit{M2OIE}         & 0.170      		&    0.090 &    0.014\\
\textit{CompactIE}         & 0.160      		&    0.093 &    0.006\\
\hline
\end{tabular}
\caption{Scores of systems on downstream tasks.} 
\label{table:ScoresDT}
\end{table}

\paragraph{\textbf{ABQA}} In \textit{ABQA}, the input is a passage of a few sentences along with a question, and the output is the answer to that question identified in the passage.  \newcite{yan2018assertionbased} created an \textit{ABQA} reference, \(\wa\) by using \textit{ClausIE} to extract all facts from the passages and asking annotators to identify  extractions answering the given question. We run all tested systems on the first 100 passages, using our  matching function to match extractions of systems to answers, and compute scores for the task by giving systems one point for a single sentence if one of their extractions match any answer cluster.

%Here again, \textit{MinIE} outperform other extractors.

\paragraph{\textbf{C-QA}} \newcite{Lu_2019} introduce a novel approach to answering complex questions called \textit{QUEST}. This system uses OIE extractions from web documents to construct a quasi Knowledge Graph and uses that graph to answer questions. They evaluate this system on a few QA benchmarks, including \textit{WikiAnswers} (CQ-W) \cite{abujabal2017automated}. We use our tested systems to extract facts from the web documents on the first 50 questions of \textit{WikiAnswers}. We then use \textit{QUEST} to construct answers and report the scores measured by the Mean Reciprocal Rank.

%Scores differ from \newcite{Lu_2019} because their scoring allowed for a single answer to contain multiple elements, which we believe to be wrong and modified.

\paragraph{\textbf{KBP}} \textit{KnowledgeNet} is a dataset of more than 7000 annotated sentences introduced by \newcite{mesquita-etal-2019-knowledgenet} for evaluating the task of automatically populating a Knowledge Base. It's annotations contain triples similar to OIE ones, with a fixed subset of 15 relations, akin to the traditional Information Extraction task. Here we measure OIE system's ability by running them on the input sentence and counting how many triples from \textit{KnowledgeNet} they were able to extract correctly.

%\begin{table}[htbp]
%\centering
%\begin{tabular}{|l|l|l|}
%\hline
%\textbf{System}         & \textbf{Score} & \textbf{Rank} \\ \hline \hline
%\textit{ReVerb}         & 0.23           & 2             \\  
%\textit{ClausIE}        & 0.18           & 3             \\  
%\textit{\textbf{MinIE}} & \textbf{0.27}  & \textbf{1}    \\ 
%\textit{IMojIE}         & 0.17           & 4             \\ 
%\textit{OpenIE6}        & 0.17           & 5             \\  
%\textit{M2OIE}          & 0.16           & 7             \\  
%\textit{CompactIE}      & 0.17           & 6             \\ \hline
%\end{tabular}
%\caption{System scores on \(\wafl\). Systems are ordered by publication date (oldest systems first).} 
%\label{table:ScoresWebAss}
%\end{table}

We compute Pearson product-moment correlation coefficients between these rankings and the ones obtained using OIE benchmarks and observe (see Table \ref{table:CorrRefsDT}) that \( \bfl \) has the highest correlation on these varied tasks, leading us to hypothesize that it is a better indicator of real performances of extractors. We also observe strong trends in the rankings (the performance on \(\bfl\) always being the best indicator, followed by those measured on \benchie, \textit{WiRE57} and finally \textit{CaRB}), further reinforcing our claim that these results should hold for most tasks.

\begin{table}[htbp]
\centering
\begin{tabular}{|l|c|c|c|}
\hline
\textbf{Benchmark} & \textbf{ABQA} & \textbf{C-QA} & \textbf{KBP} \\ \hline \hline
\textit{WiRE57}    & 0.044 & -0.184 & 0.204                                  \\  
\textit{CaRB}      & -0.649 & -0.575 & -0.382                               \\ 
\textit{BenchIE}   & 0.616  & 0.305 & 0.631                                 \\  
\( \bfl \)    & \textbf{0.940}  & \textbf{0.504} & \textbf{0.941}                      \\ \hline
\end{tabular}
\caption{Correlation between system scores on benchmarks and on downstream tasks. } 
\label{table:CorrRefsDT}
\end{table}

\subsection{Benchmark comparison}

\begin{figure}[htbp]
\center
\includegraphics[scale=0.1175]{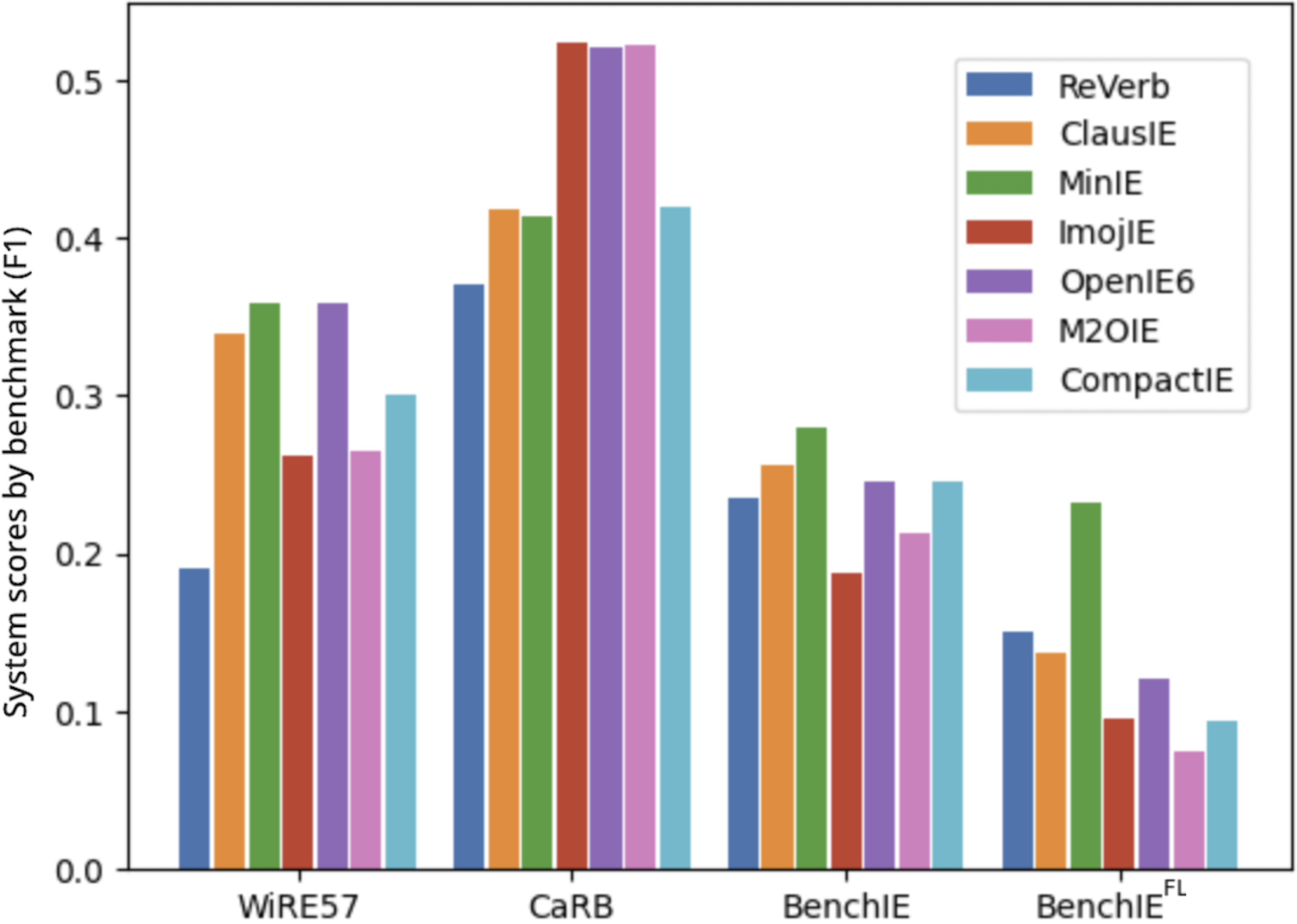}
\caption{System performance by benchmarks, scored using default scoring function of each benchmark.} 
\label{fig:sysperf}
\end{figure}

We can compare benchmarks by the different system rankings they lead to. Figure \ref{fig:sysperf} shows scores of tested systems on the four aforementioned benchmarks, using their default evaluation toolkit. 

We observe that all rankings differ largely. Certain similarities can be observed between the results obtained by \textit{BenchIE} and our reference, both of which rank \textit{MinIE} as the best-performing system. However, for virtually all other systems, there are (major) differences between their final rankings according to these two references. It is also remarkable that both recent state-of-the art benchmarks, \textit{BenchIE} and \textit{CaRB} largely overestimate neural models compared to \( \bfl \). In fact, we observe that neural networks by the virtue of the datasets they have been trained on, have a tendency of copying large chunks of input texts,\footnote{\textit{IMojIE} and \textit{M2OIE} produce tuples with an average number of words of 13.4 and 12.2 respectively, while for instance, \textit{ReVerb} produces tuples of 7.5 words on average.} that often lead to non informative tuples. See  Appendix~\ref{app:length} for an analysis of the shortcomings of neural systems.
 
% =====
% =====
\section{Discussion}
\label{sec:conclusion}

We propose new annotated resources: most notably a re-annotated OIE corpora, \( \bfl\) and a set of matching annotations, \(\bflc\). We also propose new guidelines for the OIE task, both for the annotation of tuples and the matching of candidate extractions. We also deliver noticeable improvements on the exact matching function, while not compromising the principles behind \benchie.
 
%We propose 5 new annotated resources: 2 annotated OIE corpora (\( \bfl\) and \(\wfl\)) totalizing 357 annotated sentences, 2 sets of matching annotations (\(\bflc\) and \(\wflc\)) for a total of 107 sentences and extractions produced by 7 systems, and one annotated ABQA reference \(\wafl\). We also propose new guidelines for the OIE task, both for the annotation of tuples and the matching of candidate extractions. We also deliver noticeable improvements on the exact matching function, while not compromising the principles behind \benchie.

Thanks to those resources, we conduct a meaningful comparison of off-the-shelf extractors, showing that older rule-based systems are still competitive. Finally we conduct a study that shows on three downstream tasks that our benchmark better reflects the performance of OIE systems on those tasks, making \(\bfl\) the best reference to guide OIE system choice, and to influence system development.

There are a number of avenues worth pursuing along this work, especially in annotating more sentences, both to serve as training sets for new systems and in other languages.

% =====
% =====
\section*{Limitations}

\(\bfl\) aims to correct some of the shortcomings of  \benchie. However, it still has some limitations. 

\begin{itemize}

\item While it contains a fair number of sentences to draw conclusions about which OIE systems to use, it lacks the needed size to be useful in training models. Annotation of sentences requires a lot of effort, but adding to our 300 sentences would likely be useful.

\item We believe our annotations to be more rigorous than previous ones. Still, it is likely that some mistakes were made and should be corrected. It would have been great (but out of the scope of this work) to enrol more annotators to better measure their agreement while annotating according to our guidelines. 

\item Even if the matching function we propose performs better than the exact match function of \benchie, it still lacks some flexibility. We were not able to train a better function, but some different features or methods may outperform our custom function. 

\item  Still, \(\bfl\), like most of OIE benchmarks, is only available in English.
\end{itemize}

% felipe: il faudrait glisser quelque par que l'annotation a pris bcp de temps
%We estimate that the annotations we conducted,  altogether required over 3 man-month efforts. 

%fabrice : Je crois comprendre que la section Ethics est optionnelle mais encouragée. Elle me semle pas nécéssaire ici à moins qu'on veuille discuter des biais potentiels dans les phrases\annotations utilisées qui se répercuterait dans l'évaluation des systèmes, renforcant potentiellement ces biais.

% felipe: je ne sais pas. Quel est le descriptif exactement ? Je ne pense pas que l'on doive remplir ceci. Ou alors pour souligner que notre corpus est pour l'anglais seulement ce qui favorise la segregation par le language. Mais qu'en contre-partie, nous fournissons des guides qui devrakent faciliter l'annotation dans d'autres langues.

\section*{Ethics Statement}

% Entries for the entire Anthology, followed by custom entries
\bibliography{custom}

\appendix

% =====
\clearpage
\section{Implementation details}
\label{app:implementation}

We used implementations of various systems and benchmarks in order to test and compare them to each other. 
Here is a brief description of the systems we used: 
 
\begin{description}

\item[ReVerb] 
This system uses regular expressions to extract candidate facts and a simple feature-based classifier to filter duplicates and non-useful tuples.\footnote{\url{http://reverb.cs.washington.edu/}}

\item[ClausIE]
This system uses grammatical analysis to create \textit{clauses}, or minimal sentences and extracts tuples based on those clauses and their \textit{type}. We could not find a working implementation of \textit{ClausIE} so we used a simple script to generate extraction from a web page providing an API to demonstrate the system's capabilities.\footnote{\url{https://gate.d5.mpi-inf.mpg.de/ClausIEGate/ClausIEGate/}}

\item[MinIE] This system is built on top of \textit{ClausIE}, adding more patterns to identify clauses and modifying final extractions to be minimal and have more solid relations. We used a Python wrapper of the official implementation.\footnote{\url{https://github.com/mmxgn/miniepy}}

\item[IMojIE] This system is a sequence to sequence model using a BERT encoder, trained on extractions from \textit{OpenIE4}, \textit{RnnOIE} and \textit{ClausIE}.\footnote{\url{https://github.com/dair-iitd/imojie}}

\item[OpenIE6] This system is a succesor to \textit{OllIE}, and it is a grid-labelling model built on BERT embeddings with syntaxic and grammatical constraints used at inference time trained on boot-strapped \textit{IMojIE} extractions.\footnote{\url{https://github.com/dair-iitd/openie6}}

\item[M2OIE] 
This system is one of the only multilingual system, only needing corresponding language BERT models, that works by first extracting relations and then extracting related arguments in a sequence-to-sequence architecture. \footnote{\url{https://github.com/youngbin-ro/Multi2OIE}}

\item[CompactIE]
This systems works by extracting constituents (arguments and relations) and linking them using a neural classifier, and is trained to output \textit{compact}, or minimal extractions by adding constraints on the extraction components.\footnote{\url{https://github.com/FarimaFatahi/CompactIE}}
\end{description}

All systems were ran on an Apple M2 processor. Most systems had dependencies issues out of the box and significant effort was needed to make them work as expected. Since our datasets were quite small, \textit{CaRB} being the largest with 1200 sentences, computation time was not an issue, although all neural systems needed far more time to extract facts than rule-based systems. \textit{IMojIE} was the slowest system, needing more than four hours to run on \textit{CaRB}'s test set.

Regarding the benchmarks we have been using: 

\begin{description}

\item[WiRE57] : This benchmark is comprised of a small, expertly annotated corpus of 57 sentences and uses a token-level matching function.\footnote{\url{https://github.com/rali-udem/WiRe57}}
\item[CaRB] : This benchmark which is a crowd-sourced annotation of 1200 sentences also uses a token-level scoring function. We used the test set which is made up of a total of 640 sentences.\footnote{\url{https://github.com/dair-iitd/CaRB}}
\item[BenchIE] : This benchmark uses \textit{clusters} or \textit{synsets} to group all formulations of a single fact, allowing usage of an exact matching function. It is made up of 300 manually annotated sentences. We used the default facet.\footnote{\url{https://github.com/gkiril/benchie}}

\end{description}

% =====
\section{Matching function}
\label{app:matching}
\subsection{Alternative formulations}
Here we explain the details of implementation of the alternative formulations we introduce in section \ref{sec:matching}. The aim of these alternative formulations is to match extractions to annotations even when an exact match does not show correspondence. We introduce the notion of rewriting pairs (A,B) where we authorize an argument of an extraction that contains both A and B to be rewritten by removing either A or B. We called these modified extractions \textit{alternative formulations}

We identify two cases where alternative formulations can be generated safely: in extractions with the relation \texttt{is}, and in those where an argument contains the token \texttt{and}. In the first case, we collect from the reference all pairs (A,B) from formulations (A,r,B) where r reduces to \texttt{is},\footnote{We say $r$ reduces to $r'$, if removing optional words from r leads to r'.} while in the second case, we collect pairs (A,B) whenever we have two tuples (E,rel,A) and (E,rel,B) in the reference. Table \ref{table:alterform} shows an example of how these pairs are generated from a sample annotation and extraction and how they can match annotations.
\subsection{Scoring extractions}

If a candidate extraction does not exact match a reference cluster, we consider alternative extractions obtained by rewriting pairs and give credit to the original extraction if one of those alternatives exact match the reference. We only consider the first matching alternative and its associated cluster for the computation of precision and recall as to not overly reward extractions for combining information and thus not respecting the minimality principle.

\begin{table}[htbp]
\begin{tabular}{|l|}
\hline
\textit{\begin{tabular}[c]{@{}l@{}}Chilly Gonzales is a Canadian musician who\\ lived in Paris, France and in Cologne, Germany.\end{tabular}} \\ \hline \hline 
\begin{tabular}[c]{@{}l@{}}\textbf{Annotations}:\\ (Chilly Gonzales\sep lived in\sep Paris)\\ \\ (Chilly Gonzales\sep lived in\sep Cologne)\\ \\ (Chilly Gonzales\sep is\sep Canadian)\\ \\ (Chilly Gonzales\sep is {[}a{]}\sep musician)\end{tabular} \\ \hline
\begin{tabular}[c]{@{}l@{}}\textbf{Rewriting pairs}:\\ \texttt{is} : (Canadian\sep  Musician)\\ \texttt{and} : (Paris\sep Cologne)\end{tabular}                                     \\ \hline
\begin{tabular}[c]{@{}l@{}}\textbf{Extraction:}\\ (Chilly Gonzales\sep is a\sep Canadian musician)\\ \\ \textbf{Alternative formulations:}\\ (Chilly Gonzales\sep is\sep Canadian)\\ (Chilly Gonzales\sep is\sep musician)\end{tabular} \\ \hline
\begin{tabular}[c]{@{}l@{}}\textbf{Extraction:}\\ (Chilly Gonzales\sep lived in\sep Paris and Cologne)\\ \\ \textbf{Alternative formulations:}\\ (Chilly Gonzales\sep lived in\sep Paris)\\ (Chilly Gonzales\sep lived in\sep Cologne)\end{tabular} \\ \hline
\end{tabular}
\caption{Alternative formulation generation example} 
\label{table:alterform}
\end{table}

% =====
\section{About inferred clusters}
\label{app:infereds}

Since our annotation and \textit{BenchIE}’s differ largely in handling of inference, we present a modified version of our annotation set for which the original annotator transformed the inferred clusters to use only words present in the original sentence (light inference) and removed the inferred clusters that could not be modified as such (heavy inference). This modified annotation is what is presented to the annotator when comparing ours and \textit{BenchIE}'s annotations in section \ref{sec:manEval}, in order to give a fair comparison in regards to exhaustiveness given that \textit{BenchIE}'s annotation does not contain inference, other than in some very rare cases. Statistics regarding this modification and the use of inference in both references are presented in Table \ref{table:caracteristiquesinf}. We observe that in our original annotation, we had 22 annotated clusters that were instances of heavy inference which we were not able to transform into light inference and had to be removed, while 22 of them could be transformed. We also observe that \textit{BenchIE} has very few instances of inference. This is because they decide to only annotate facts for which the relation is verbatim in the text, although we did find a few instances of light inference.

%\felipe{quite hard to understand. We must analyze this table anyway}{Still the problem of inference which is not clearly presented, and we don't even describe what benchie does regarding this (we have to guess this is like second column)}. 

\begin{table}[htbp]
\resizebox{\columnwidth}{!}{%
\begin{tabular}{l|l|l|l|}
\cline{2-4}
\multicolumn{1}{c|}{\textbf{}} &
  \multicolumn{1}{c|}{{\color[HTML]{000000} \textbf{\(\bfl\)}}} &
  \multicolumn{1}{c|}{\textbf{\begin{tabular}[c]{@{}c@{}}Modified\end{tabular}}} &
  \textbf{\textit{BenchIE}} \\   \hline 
\multicolumn{1}{|l|}{Total clusters}    & 139       & 117       & 119       \\ 
\multicolumn{1}{|l|}{Inferred clusters} & 47 (34\%) & 25 (21\%) & 3 (2.5\%) \\  
\multicolumn{1}{|l|}{Heavy inference}   & 44 (32\%) & 0         & 0         \\  
\multicolumn{1}{|l|}{Light inference}   & 3 (2\%)   & 25 (21\%) & 3 (2.5\%) \\ \hline
\end{tabular}%
}
\caption{Inference statistics} 
\label{table:caracteristiquesinf}
\end{table}

% =====
\section{Comparison of scoring functions}
\label{app:comparison_sf}

Using our manual matching annotations, we compute theoretical scores and rankings for tested systems. We then compute scores and rankings using the different matching scores and functions of previous benchmarks. We compare these matching methods by computing their Pearson product-moment correlation coefficients with the manual rankings on \( \bflc \) (See Table \ref{table:corrfuncanno}). We observe that token-level matching has very low correlation and that while both \textit{BenchIE}’s exact match and our \textit{Custom Match} have similar scores, ours correlates to a greater degree.

\begin{table}[htbp]
\centering
\begin{tabular}{|l|c|}
\hline
\multicolumn{1}{|c|}{Matching function} & \begin{tabular}[c]{@{}c@{}}Correlation with\\ manual rankings\end{tabular} \\ \hline \hline
\textit{WiRE57}                                           & 0.219                                                                                                              \\  
\textit{BenchIE}                                          & 0.961                                                                                                              \\
\textit{Custom Match}                                     & \textbf{0.997}                                                                                                              \\ \hline
\end{tabular}
\caption{Pearson product-moment correlation between rankings obtained by matching functions and manual rankings on \( \bflc \)} 
\label{table:corrfuncanno}
\end{table}

% =====

% =====
\begin{table*}[htbp]
\centering
\begin{tabular}{|ll|}
\hline
\multicolumn{2}{|l|}{\parbox{15cm}{\textbf{passage:} \hspace*{1.5cm} the internal frame backpack is a recent innovation , invented in 1967 by greg lowe, who went on to found lowe alpine and lowepro, companies specializing in backpacks and other forms of carrying bags for various equipment}}\\
\multicolumn{2}{|l|}{\parbox{15cm}{\textbf{question:}   \hspace*{1.18cm} when was the backpack  invented ?}}\\
\hline
\hline
\(\wa\) & (a recent innovation \sep invented \sep in 1967 \sep  by greg lowe)\\
\(\wafl\)                          &  ([the] [internal frame] backpack \sep [was]/[be]/[is] invented in \sep 1967)\\           \hline                         
\end{tabular}

\caption{Question-Passage-Answer triples from \(\wa\) and \(\wafl\).} 
\label{table:webassex}
\end{table*}
\section{Neural system scores}
\label{app:length}

Here we try to demonstrate why neural approaches get lower scores on our benchmark than on others. First we compare lengths of extractions by systems in Table \ref{table:ExtLength}, were we see that neural approaches all have longer mean extraction length compared to rule-based systems, except for \textit{CompactIE} wich has been specifically trained to output compact extractions. We also see that our two best performing systems, \textit{MinIE} and \textit{ReVerb} have the shortest extractions. We hypothesize that they obtain the highest scores in part because they are most capable at precisely separating facts into minimal extractions, which is something that neural systems fail to do. This is illustrated in Table \ref{table:ExtEx} were we show all extractions for all systems on a given sentence. We see that both \textit{IMojIE} and \textit{M2OIE}, neural systems, make extremely long and useless extractions, recopying almost the whole sentence. Indeed, we observed that neural systems have a tendency to recopy large parts of the input sentences without being able to accurately separate facts. We hypothesize that this is because they have been trained on non-minimal and relation-complete extractions.

\begin{table}[htbp]
\begin{tabular}{|l|c|}
\hline
\multicolumn{1}{|c|}{\textbf{System}} & \textbf{Mean extraction length} (\# words) \\ \hline \hline
ReVerb                       & 7.5                               \\ \hline
ClausIE                      & 10.9                              \\ \hline
MinIE                        & 7.9                               \\ \hline
IMojIE                       & 13.4                              \\ \hline
OpenIE6                      & 12.2                              \\ \hline
M2OIE                        & 12.2                              \\ \hline
CompactIE                    & 9.3                               \\ \hline
\end{tabular}
\caption{Extraction length by system} 
\label{table:ExtLength}
\end{table}

% ====

\section{Downstream Tasks Details}
\label{app:pearson}
%C-QA and KBP Scoring
Here we provide more details on how we ran the experiment on the downstream tasks and on some modifications that were made to either their data or their scoring procedures.

\paragraph{ABQA} Table~\ref{table:webassex} shows an example of a triple from \wa. We found many  tuples in it  that  are  too long or even fail to answer the question. Thus, we   re-annotated the 100 first passage-question pairs of \(\wa\), following the \benchie{} format and listing all clusters that can answer the question. We distribute the resulting resource named \(\wafl\), an example of which being in Table~\ref{table:webassex}. We believe that this did not impact the results of system on the task as we simply corrected the mistakes and formatted the answers in \benchie 's cluster format. We also only used the first 100 sentences of \wa because of the high annotation effort required.

\paragraph{C-QA} For this task, our results are far below what the original paper's scores were. This is because we slightly modified the scoring used : \textit{QUEST}'s original code allowed for various formulations of an entity to be regrouped in a single answer. However, in practice, multiple different entities were regrouped in a single answer, giving full points to answers containing multiple different elements. Using the original scoring, we found system's score to be closer to the original papers (0.25 to 0.29). We thus separated all entities into distinct answers and computed MRR on those answers, which we believe to be a more accurate evaluation method. Table \ref{table:wikianswersex} shows an example of a single question from \textit{WikiAnswers} along with the answer modification procedure.

\begin{table*}[htbp]
\resizebox{\textwidth}{!}{%
\begin{tabular}{|l|l|}
\hline
Question                & which films were starred by julia roberts and richard gere?                                                                                                                                                                                                                                                                                                                                                                        \\ \hline
Gold Answer             & runaway bride, pretty woman                                                                                                                                                                                                                                                                                                                                                                                                        \\ \hline
Documents               & \begin{tabular}[c]{@{}l@{}}Document 1 :\\ In his career Richard Gere has starred opposite some seriously attractive women and ...\\ With Kim Basinger he starred in 2 movies "No Mercy" and "Final Analysis", with Diane Lane...\\ ...\\ Document 2 :\\ Julia Fiona Roberts is an American actress and producer who made her debut in ...\\ Roberts made her breakthrough the following year by starring in ...\\ ...\end{tabular} \\ \hline
Original QUEST Answer   & \begin{tabular}[c]{@{}l@{}}Answer 1 : ('a year', 'golden globe', 'pretty woman', 'jason alexander' )\\ Answer 2 : ('phillip', '41', 'richard')\\ ...\\ MRR : 1\end{tabular}                                                                                                                                                                                                                                                        \\ \hline
Reformated QUEST Answer & \begin{tabular}[c]{@{}l@{}}Answer 1 : 'a year'\\ Answer 2 : 'golden globe'\\ Answer 3 : 'pretty woman'\\ ...\\ MRR  : 0.333\end{tabular}                                                                                                                                                                                                                                                                                           \\ \hline
\end{tabular}%
}
\caption{\textit{WikiAnswers} Example with output modification.} 
\label{table:wikianswersex}
\end{table*}

\textit{IMojIE} is not included in this experiment because of compute limitations. Indeed, \textit{QUEST} uses 10 documents per questions, each document containing a few hundred sentences, making the total sentence count for 50 questions more than 50,000. \textit{IMojIE} being the slowest system, running it on this whole corpus would have taken more than two months of compute time on our setup.

\paragraph{KBP} Table \ref{table:knowEx} shows an example of annotation from \textit{KnowledgeNet}. This reference's annotated relations come from a fixed subset of 15 relations, like \texttt{PlaceOfBirth} or \texttt{FoundedBy}. Consequently, they do not annotate the relation words from the sentence, making it harder to compare to our system's annotation. In order to find matches between extractions and annotations, we only compared arguments, and assumed that if both arguments were absolutely equal to annotations, the information should be very similar. This might not be perfectly accurate but does not penalise or reward systems differently and accurately measures systems ability to properly extract arguments. Furthermore, we only counted one annotation for each different relation per sentence, since the annotations in \textit{KnowledgeNet} often contained more than one annotation with the same relation for the same fact, often listing pronouns and coreference to a single entity in different annotations. 

Here, \textit{IMojIE} is once again excluded from the experiment, because of the same compute limitations. \textit{KnowledgeNet}'s corpus contains more than 7000 sentences, and running this system would have necessitated more than 10 days of compute time on our setup.

\begin{table}[htbp]
\resizebox{\columnwidth}{!}{%
\begin{tabular}{|l|}
\hline
\begin{tabular}[c]{@{}l@{}}\textbf{Sentence :} \textit{After moving to New York, where she established}\\  \textit{Euro Capital Properties along with her husband Jacques,} \\ \textit{her discerning eye for design was put to great use.}\end{tabular} \\ \hline
\begin{tabular}[c]{@{}l@{}}\textbf{Annotations :}\\ she - \texttt{PlaceOfResidence} - New York\\ her - \texttt{PlaceOfResidence} - New York\\ \\ she - \texttt{Spouse} - Jacques\\ her - \texttt{Spouse} - Jacques\end{tabular}                                \\ \hline
\end{tabular}%
}
\caption{\textit{KnowledgeNet} Annotation Example.} 
\label{table:knowEx}
\end{table}

\section{Proxys}
\label{app:proxys}
Here, we validate our re-annotation by devising objective proxies that highlight differences between \benchie 's annotations and ours. These proxies try to identify annotations from both sets that might fall into one of the error categories from our experiment in Section \ref{sec:annProb}.

\paragraph{Double Annotation} We find distinct clusters that have identical formulations.
\paragraph{Double Meaning} We find clusters that have two different meaning by selecting those that have identical first argument and relation but with different second argument.
\paragraph{False/Irrelevant/Missing Facts} We find clusters from one set of annotation that don't appear in the other set.

Table \ref{table:proxies} shows the number of sentence that have at least one cluster identified by each proxy. We first observe that all proxies return a lot more occurrences from \benchie 's annotation than from ours. This leads us to believe that this annotation contains more errors. However, not all annotations matched by the proxies are necessarily an error. Examples of such annotations from both sets are presented in Table \ref{table:proxEx}.

\paragraph{} For the \textit{Double Annotation} proxy, we find that annotations from \benchie{} are actually errors, the same fact with a differently formulated relation present in two distinct clusters. In contrast, the annotations from \(\bfl\) are cases were we find both the annotation regrouping the information in a single cluster or separating it in two would be adequate. We verify that this is the case for the 5 \textit{Double annotation} clusters from \(\bfl\), and that most observed examples from the 135 cases from \benchie{} are all double annotation errors.

\paragraph{} The same is true for the \textit{Double Meaning} proxy, where most cases from \benchie{} are errors, and almost none from \(\bfl\) are. \benchie 's example in Table \ref{table:proxEx} is especially strong, where five different meanings are present in a single cluster. However, we estimate that around 50\% of annotations matched by that proxy from \textit{BenchIE} are errors, still a very high count, and that most from \(\bfl\) are simply reformulation of the same entity or fact, like \textit{Doctor} and \textit{Dr.} from the example.

\paragraph{} For the  \textit{False/Irrelevant/Missing Facts} proxy, it is important to note that counts are inverted : the proxy returns 17 annotations from \textit{BenchIE} that are not present in \(\bfl\), and 68 from the inverse, leading us to believe that our annotation is more exhaustive. Looking at the example, we once again find that the fact present in \textit{BenchIE} and not in \(\bfl\) is actually an irrelevant fact, and that the inverse is a simple case of inference, not present in \textit{BenchIE}. We estimate that 90\% of facts present in our annotation and not in \benchie 's are actually \textit{Missing Facts}, and that more than 50\% of facts present in \textit{BenchIE} and not in our annotation are either \textit{Irrelevant} or \textit{False}.

\begin{table}[htbp]
\resizebox{\columnwidth}{!}{%
\begin{tabular}{l|ll|ll|}

\cline{2-5}
                                                    & \multicolumn{2}{l|}{\textbf{BenchIE}} & \multicolumn{2}{l|}{\textbf{\(\bfl\)}} \\ \hline
\multicolumn{1}{|l|}{Double annotation}             & \multicolumn{2}{l|}{135}       & \multicolumn{2}{l|}{5}       \\ \hline
\multicolumn{1}{|l|}{Double meaning}                & \multicolumn{2}{l|}{55}        & \multicolumn{2}{l|}{26}      \\ \hline
\multicolumn{1}{|l|}{False/Irrelevant/Missing fact} & \multicolumn{2}{l|}{17}        & \multicolumn{2}{l|}{68}      \\ \hline

\end{tabular}%
}
\caption{Proxies sentence count} 
\label{table:proxies}
\end{table}

\begin{table*}[htbp]
\begin{tabular}{|l|}
\hline
\multicolumn{1}{|c|}{\textbf{Double Annotation}}                                                                                                                                                                                                                                                                                                                                                                                                                                                                                                                                                                                             \\ \hline
\begin{tabular}[c]{@{}l@{}}\textit{BenchIE} : \textit{He served as the first Prime Minister of Australia and became}\\ \textit{a founding justice of the High Court of Australia .}\\ \\ He - became - {[}a{]} {[}founding{]} justice of {[}the{]} High Court {[}of Australia{]}\\ \\ He - became {[}a{]} {[}founding{]} justice of - {[}the{]} High Court {[}of Australia{]}\\ He - became {[}a{]} {[}founding{]} justice - of {[}the{]} High Court {[}of Australia{]}\end{tabular}                                                                                                                                                                                    \\ \hline
\begin{tabular}[c]{@{}l@{}}\(\bfl\) : \textit{For patients who do not recover quickly ,}\\ \textit{the protocol also includes support groups and/or psychotherapy .}\\ \\ {[}the{]} protocol - includes - support groups\\ {[}the{]} protocol - includes - support groups and/or psychotherapy\\ \\ {[}the{]} protocol - includes - psychotherapy\\ {[}the{]} protocol - includes - support groups and/or psychotherapy\end{tabular}                                                                                                                                                                                                                          \\ \hline
\multicolumn{1}{|c|}{\textbf{Double Meaning}}                                                                                                                                                                                                                                                                                                                                                                                                                                                                                                                                                                                                \\ \hline
\begin{tabular}[c]{@{}l@{}}\textit{BenchIE} : \textit{It deals with cases of fraud in relation to direct taxes and indirect taxes ,}\\ \textit{tax credits , drug smuggling , and money laundering , }\\ \textit{cases involving United Nations trade sanctions , conflict diamonds and CITES .}\\ \\ It - deals - with cases of fraud in {[}relation to{]} money laundering\\ It - deals - with cases of fraud in {[}relation to{]} drug smuggling\\ It - deals - with cases of fraud in {[}relation to{]} indirect taxes\\ It - deals - with cases of fraud in {[}relation to{]} direct taxes\\ It - deals - with cases of fraud in {[}relation to{]} tax credits\end{tabular} \\ \hline
\begin{tabular}[c]{@{}l@{}}\(\bfl\) : \textit{Dr. Jagan himself was personally involved in the organization of the strike ,}\\ \textit{and helped to raise funds across the country to it .}\\ \\ Jagan - is {[}a{]} - Dr.\\ Jagan - is {[}a{]} - Doctor\end{tabular}                                                                                                                                                                                                                                                                                                                                                                                         \\ \hline
\multicolumn{1}{|c|}{\textbf{False/Irrelevant/Missing Facts}}                                                                                                                                                                                                                                                                                                                                                                                                                                                                                                                                                                                \\ \hline
\begin{tabular}[c]{@{}l@{}}\textit{BenchIE} : \textit{Graner handcuffed him to the bars of a cell window and left him there ,}\\ \textit{feet dangling off the floor , for nearly five hours .}\\ \\ feet - dangling off {[}the{]} floor for - nearly five hours\end{tabular}                                                                                                                                                                                                                                                                                                                                                                                           \\ \hline
\begin{tabular}[c]{@{}l@{}}\(\bfl\) : \textit{He served as the first Prime Minister of Australia and became}\\ \textit{a founding justice of the High Court of Australia .}\\ \\ Australia - has {[}had{]} - {[}a{]} High Court\end{tabular}                                                                                                                                                                                                                                                                                                                                                                                                                  \\ \hline
\end{tabular}
\caption{Examples of proxy outputs} 
\label{table:proxEx}
\end{table*}

\section{Examples of Extractions}
Table \ref{table:ExtEx} shows example of extractions from all 7 tested systems on a single sentence.
\begin{table*}[htbp]
\resizebox{\textwidth}{!}{%
\begin{tabular}{|ll|}
\hline
\multicolumn{2}{|l|}{\begin{tabular}[c]{@{}l@{}}His son , John Crozier , Jr. , was an early aviation pioneer who began building a human-powered\\  flying machine in the 1890s , but was killed in a feud in Grainger County before he could complete it .\end{tabular}} \\ \hline
\multicolumn{1}{|c|}{\textbf{System}} &
  \multicolumn{1}{c|}{\textbf{Extractions}} \\ \hline \hline
\multicolumn{1}{|l|}{ReVerb} &
  \begin{tabular}[c]{@{}l@{}}(John Crozier\sep was\sep an early aviation pioneer)\\ (an early aviation pioneer\sep began building\sep a human-powered flying machine)\\ (the 1890s\sep was killed in\sep a feud)\\ (he\sep could complete\sep it)\end{tabular} \\ \hline
\multicolumn{1}{|l|}{ClausIE} &
  \begin{tabular}[c]{@{}l@{}}(His\sep has\sep son)\\ (His son\sep is\sep John Crozier)\\ (John Crozier\sep is\sep Jr.)\\ (His son\sep was\sep an early aviation pioneer)\\ (an early aviation pioneer\sep began\sep\\ building a human-powered flying machine in the 1890s)\\ (an early aviation pioneer\sep began\sep building a human-powered flying machine)\\ (an early aviation pioneer\sep was killed\sep\\ in a feud in Grainger County before he could complete it)\\ (an early aviation pioneer\sep was killed\sep in a feud in Grainger County)\\ (he\sep could complete\sep it)\end{tabular} \\ \hline
\multicolumn{1}{|l|}{MinIE} &
  \begin{tabular}[c]{@{}l@{}}(His\sep has\sep son)\\ (son\sep is\sep John Crozier)\\ (John Crozier\sep is\sep son)\\ (John Crozier\sep is\sep Jr.)\\ (son\sep was\sep early aviation pioneer)\\ (early aviation pioneer\sep began building human-powered flying machine in\sep\\ the 1890s)\\ (early aviation pioneer\sep began\sep building human-powered flying machine)\\ (early aviation pioneer\sep was killed in feud in\sep Grainger County)\\ (he\sep complete\sep it)\end{tabular} \\ \hline
\multicolumn{1}{|l|}{IMojIE} &
  \begin{tabular}[c]{@{}l@{}}\textbf{(His son}\sep \textbf{was}\sep\\ \textbf{an early aviation pioneer who began building a human-powered flying} \textbf{machine}\\ \textbf{in the 1890s , but was killed in a feud in Grainger County)}\\ (he\sep could complete\sep it)\end{tabular} \\ \hline
\multicolumn{1}{|l|}{OpenIE6} &
  \begin{tabular}[c]{@{}l@{}}(His son\sep was\sep\\ an early aviation pioneer who began building a human-powered flying machine in the 1890s)\\ (an early aviation pioneer\sep began\sep\\ building a human-powered flying machine in the 1890s)\\ (an early aviation pioneer\sep was killed\sep\\ in a feud in Grainger County before he could complete it)\\ (he\sep could complete\sep it)\\ (an early aviation pioneer\sep began building\sep a human-powered flying machine in the 1890s)\end{tabular} \\ \hline
\multicolumn{1}{|l|}{M2OIE} &
  \begin{tabular}[c]{@{}l@{}}\textbf{(His son , John Crozier , Jr.}\sep \textbf{was}\sep\\\textbf{ an early aviation pioneer who began building a human-}\textbf{powered flying machine in the 1890s ,} \\ \textbf{but was killed in a feud in Grainger County)}\\ (an early aviation pioneer\sep began building\sep a human-powered flying machine in the 1890s)\\ (an early aviation pioneer\sep was killed\sep in a feud in Grainger County)\\ (he\sep could complete\sep it)\end{tabular} \\ \hline
\multicolumn{1}{|l|}{CompactIE} &
  \begin{tabular}[c]{@{}l@{}}(John Crozier\sep was\sep an early aviation pioneer)\\ (an early aviation\sep was killed\sep in a feud in Grainger County)\\ (an early aviation\sep could complete\sep it)\end{tabular} \\ \hline
\end{tabular}%
}
\caption{Extraction examples} 
\label{table:ExtEx}
\end{table*}
\section{Annotation Examples}
\label{sec:appendixanno}
Table \ref{table:AnnoEx} shows differences between our annotation and \textit{BenchIE}'s original annotations for the same sentences. We see that our annotation has a lot more clusters, partly because of our inclusion of inference but also because of the minimality principle that guides us to annotate the fact (My Classical Way \sep was \sep released), which is not annotated in \textit{BenchIE}. We also see that \textit{BenchIE}'s annotation contain a lot of different formulations of the same fact, most of which we do not consider to be valid because of relation integrity.
\begin{table*}[htbp]
\centering
\resizebox{\textwidth}{!}{%
\begin{tabular}{|cl|}
\hline
\multicolumn{2}{|c|}{\textit{`` My Classical Way '' was released on 21 September 2010 on Marc 's own label , Frazzy Frog Music .}} \\ \hline
\multicolumn{1}{|c|}{\textbf{BenchIE's annotatios}} &
  \multicolumn{1}{c|}{\textbf{\( \bfl\)'s annotations}} \\ \hline
\multicolumn{1}{|l|}{\begin{tabular}[c]{@{}l@{}}Cluster 1:\\ {[}` `{]} My Classical Way {[}' '{]} \sep was released on \sep 21 September 2010\\ {[}` `{]} My Classical Way {[}' '{]} \sep was \sep released on 21 September 2010\\ {[}` `{]} My Classical Way {[}' '{]} \sep was released \sep on 21 September 2010\\ {[}``{]} My Classical Way {[}''{]} \sep was released on \sep 21 September 2010\\ {[}``{]} My Classical Way {[}''{]} \sep was \sep released on 21 September 2010\\ {[}``{]} My Classical Way {[}''{]} \sep was released \sep on 21 September 2010\\ ``My Classical Way'' \sep was released on \sep 21 September 2010\\ ``My Classical Way'' \sep was \sep released on 21 September 2010\\ ``My Classical Way'' \sep was released \sep on 21 September 2010\\ \\ Cluster 2:\\ {[}` `{]} My Classical Way {[}' '{]} \sep was released on \sep Marc 's {[}own{]} label\\ {[}` `{]} My Classical Way {[}' '{]} \sep was \sep released on Marc 's {[}own{]} label\\ {[}` `{]} My Classical Way {[}' '{]} \sep was released \sep on Marc 's {[}own{]} label\\ {[}` `{]} My Classical Way {[}' '{]} \sep was released on \sep Frazzy Frog Music\\ {[}` `{]} My Classical Way {[}' '{]} \sep was \sep released on Frazzy Frog Music\\ {[}` `{]} My Classical Way {[}' '{]} \sep was released \sep on Frazzy Frog Music\\ {[}``{]} My Classical Way {[}''{]} \sep was released on \sep Marc 's {[}own{]} label\\ {[}``{]} My Classical Way {[}''{]} \sep was \sep released on Marc 's {[}own{]} label\\ {[}``{]} My Classical Way {[}''{]} \sep was released \sep on Marc 's {[}own{]} label\\ {[}``{]} My Classical Way {[}''{]} \sep was released on \sep Frazzy Frog Music\\ {[}``{]} My Classical Way {[}''{]} \sep was \sep released on Frazzy Frog Music\\ {[}``{]} My Classical Way {[}''{]} \sep was released \sep on Frazzy Frog Music\\ ``My Classical Way'' \sep was released on \sep Marc 's {[}own{]} label\\ ``My Classical Way'' \sep was \sep released on Marc 's {[}own{]} label\\ ``My Classical Way'' \sep was released \sep on Marc 's {[}own{]} label\\ ``My Classical Way'' \sep was released on \sep Frazzy Frog Music\\ ``My Classical Way'' \sep was \sep released on Frazzy Frog Music\\ ``My Classical Way'' \sep was released \sep on Frazzy Frog Music\end{tabular}} &
  \begin{tabular}[c]{@{}l@{}}Cluster 1:\\ {[}`{]} {[}`{]} My Classical Way {[}''{]} \sep was \sep released\\ \\ Cluster 2:\\ {[}`{]} {[}`{]} My Classical Way {[}''{]} \sep was released on \sep 21 September 2010\\ \\ Cluster 3:\\ {[}`{]} {[}`{]} My Classical Way {[}''{]} \sep was released on \sep Frazzy Frog Music\\ \\ Cluster 4:\\ Frazzy Frog Music \sep is \sep Marc {[}'s{]} own label\\ Frazzy Frog Music \sep is own label of \sep Marc\\ Frazzy Frog Music \sep is owned by \sep Marc\\ Marc {[}'s{]} own label \sep is \sep Frazzy Frog Music\\ \\ Cluster 5:\\ Frazzy Frog Music \sep is {[}a{]} \sep label\\ Frazzy Frog Music \sep is \sep {[}a{]} label\\ \\ Cluster 6:\\ Marc \sep has {[}a{]} \sep label\\ Marc \sep has \sep {[}a{]} label\\ Marc \sep has \sep {[}own{]} label\\ Marc \sep owns {[}a{]} \sep label\\ Marc \sep owns \sep {[}a{]} label\end{tabular} \\ \hline
\end{tabular}%
}
\caption{Annotations examples from \textit{BenchIE} and \(\bfl\)} 
\label{table:AnnoEx}
\end{table*}

% =====
\clearpage
\section{Annotation Guidelines}
\label{sec:appendixannoG}

This Appendix contains annotation guidelines for the open information extraction task and has been used in the annotation process for the \(\bfl\) reference. The various principles dictate which facts should and should not be annotated. The information is presented in the following format: Sentences are in the top cell and in the cell bellow, examples that should be annotated are in \textcolor{fecorrect}{green} and preceded by a check mark, ones that should not be included are in \textcolor{fewrong}{red} and preceded by a cross mark.
\subsection{Number of arguments}
All tuples must contain between 1 and 2 arguments. Extractions with more than two arguments can be split into more compact extractions. The first example shows this principle, while the second example shows how some tuples only have a single argument (we write XXX in the second argument for convenience).
\begin{table}[htbp]
\centering
\begin{tabular}{|l|}
\hline
\textit{Kyle left for school on Monday.}                                                                                                \\ \hline
\begin{tabular}[c]{@{}l@{}}\wrong{(Kyle \sep left \sep for school \sep on Monday)}\\ \\ \correct{(Kyle \sep left on \sep Monday)}\\ \\ \correct{(Kyle \sep left for \sep school)}\end{tabular} \\ \hline
\end{tabular}
\caption*{Number of arguments : First example} 
\label{table:AnnoNumArgs1}
\end{table}

\begin{table}[htbp]
\centering
\begin{tabular}{|l|}
\hline
\textit{\begin{tabular}[c]{@{}l@{}}Gideon Rodan taught at the University\\  of Connecticut School of Dental Medicine.\end{tabular}} \\ \hline
\begin{tabular}[c]{@{}l@{}}\correct{(Gideon Rodan \sep taught \sep XXX)}\\ \correct{(Gideon Rodan \sep was \sep a teacher)}\end{tabular}                                \\ \hline
\end{tabular}
\caption*{Number of arguments : Second example} 
\label{table:AnnoNumArgs2}
\end{table}
\subsection{Informativeness}
Annotated tuples must contain relevant information that is expressed in the sentence. Tuples must be informative and relevant. They must not contain generality or empty words that convey no information. In the example, the fact that \textit{he has written} is not relevant since it is a generality, most people \textit{have written} and it is not the information presented in the sentence.
\begin{table}[htbp]
\centering
\begin{tabular}{|l|}
\hline
\textit{\begin{tabular}[c]{@{}l@{}}He has written several newspaper and\\  magazine opinion pieces in The Guardian,\\  Vice, Billboard, and others.\end{tabular}} \\ \hline
\begin{tabular}[c]{@{}l@{}}\correct{(He \sep has written \sep opinion pieces)}\\ \\ \wrong{(He \sep has \sep written)}\end{tabular}                                                                 \\ \hline
\end{tabular}
\caption*{Informativeness} 
\label{table:AnnoInf}
\end{table}

\subsection{Minimality}
Annotated tuples must contain minimal information, which cannot be subdivided into smaller pieces of information. No argument should contain information about two different entities if this is true for both, and no tuple should contain more than one piece of information about an entity if these can be divided. In the first example, all the different minimal pieces of information (creators, time of creation) must be separated in minimal clusters and not grouped like in the example that should not be included.
\begin{table}[htbp]
\centering
\resizebox{\columnwidth}{!}{%
\begin{tabular}{|l|}
\hline
\textit{The group was created in 2020 by three people} \\ \hline
\begin{tabular}[c]{@{}l@{}}\correct{(The group \sep was \sep created)}\\ \\ \correct{(The group \sep was created in \sep 2020)}\\ \\ \correct{(The group \sep was created by \sep three people)}\\ \\ \wrong{(The group \sep was \sep} \\\textcolor{fewrong}{created in 2020 by three people)}\end{tabular} \\ \hline
\end{tabular}%
}
\caption*{Minimality : First example} 
\label{table:AnnoMin1}
\end{table}
It is sometimes necessary to separate information, if and only if it is also true when separated. In the second example, the dog is neither \textit{black} nor \textit{brown}, but \textit{black} \textbf{and} \textit{brown}, whereas in the first example, \textit{He has Cornish ancestors} \textbf{and} \textit{He has Welsh ancestors}.

\begin{table}[htbp]
\centering
\begin{tabular}{|l|}
\hline
\textit{He has Cornish as well as Welsh ancestry.}                                                 \\ \hline
\begin{tabular}[c]{@{}l@{}}\correct{(He \sep has \sep Cornish ancestry)}\\ \\ \correct{(He \sep has \sep Welsh ancestry)}\end{tabular} \\ \hline
\end{tabular}
\caption*{Minimality : Second example} 
\label{table:AnnoMin2}
\end{table}

\begin{table}[htbp]
\centering
\begin{tabular}{|l|}
\hline
\textit{The dog is black and brown.} \\ \hline
\correct{(The dog \sep is \sep black and brown)}       \\ \hline
\end{tabular}
\caption*{Minimality : Third example} 
\label{table:AnnoMin3}
\end{table}

\subsection{Exhaustivity}
All the minimal information present in the sentence must be included in the annotations. Some arguments or relations may be affected by modifications but remain true without them, so it's necessary to list all possible formulations that respect the other principles. In the example, it is true that \textit{he wrote opinion pieces, newspaper opinion pieces} and \textit{magazine opinion pieces}, so all these facts must be listed in three separate clusters.
\begin{table}[htbp]
\centering
\begin{tabular}{|l|}
\hline
\textit{\begin{tabular}[c]{@{}l@{}}He has written several newspaper and\\  magazine opinion pieces in The Guardian,\\  Vice, Billboard, and others.\end{tabular}} \\ \hline
\begin{tabular}[c]{@{}l@{}}\correct{(He \sep has written \sep }\\\textcolor{fecorrect}{several newspaper opinion pieces)}\\ \\ \correct{(He \sep has written \sep }\\\textcolor{fecorrect}{several magazine opinion pieces)}\\ \\ \correct{(He \sep has written \sep }\\\textcolor{fecorrect}{several opinion pieces)}\end{tabular} \\ \hline
\end{tabular}

\caption*{Exhaustivity} 
\label{table:AnnoExhau}
\end{table}

\subsection{Relation completeness}
Relations are the vehicles of information; arguments must not contain information that changes their meaning. Relations can be complicated but necessary, while they can sometimes be simplified. They must be simplified as much as possible to respect the principle of minimality, without losing their original meaning, expressed in the sentence. In the example, the second argument of the erroneous annotation contains the word \textit{over}, which modifies the meaning of the relation \textit{is}, whereas in the positive example, the second argument, \textit{13 Millions}, is only the object of the relation.

\begin{table}[htbp]
\centering
\begin{tabular}{|l|}
\hline
\textit{Tokyo’s population is over 13 Millions}                                                                                    \\ \hline
\begin{tabular}[c]{@{}l@{}}\correct{(Tokyo’s population \sep \textbf{is over} \sep 13 Millions)}\\ \\ \wrong{(Tokyo’s population \sep \textbf{is} \sep over 13 Millions)}\end{tabular} \\ \hline
\end{tabular}
\caption*{Relation completeness : First example} 
\label{table:AnnoRelComp1}
\end{table}

Sometimes, relationships can be complicated but necessary, while sometimes they can be simplified, keeping the additional part optional only if it's made necessary by the lack of other tuples explaining that additional part, as in the second example where the part \textit{from Hungary} in the second cluster is optional because the place of origin of their escape is present in the first cluster. In the third example, \textit{in Paris} is not optional because without this information, the relation no longer holds.

\begin{table}[htbp]
\centering
\begin{tabular}{|l|}
\hline
\textit{\begin{tabular}[c]{@{}l@{}}His parents are Ashkenazi Jews who had to\\  flee from Hungary during World War II.\end{tabular}}                     \\ \hline
\begin{tabular}[c]{@{}l@{}}\correct{(His parents \sep had to flee from \sep Hungary)}\\ \\ \correct{(His parents \sep }\\\textcolor{fecorrect}{had to flee \textbf{{[}from Hungary{]}} during \sep} \\\textcolor{fecorrect}{World War II)}\end{tabular} \\ \hline
\end{tabular}
\caption*{Relation completeness : Second example} 
\label{table:AnnoRelComp2}
\end{table}

\begin{table}[htbp]
\centering
\resizebox{\columnwidth}{!}{%
\begin{tabular}{|l|}
\hline
\textit{\begin{tabular}[c]{@{}l@{}}Chilly Gonzales is a Grammy-winning Canadian\\  musician who resided in Paris, France for several\\  years, and now lives in Cologne, Germany.\end{tabular}} \\ \hline
\correct{(Chilly Gonzales \sep} \\\textcolor{fecorrect}{resided \textbf{in Paris} for \sep }\\\textcolor{fecorrect}{several years)} \\ \hline
\end{tabular}%
}
\caption*{Relation completeness : Third example} 
\label{table:AnnoRelComp3}
\end{table}

\subsection{Coreference resolution}
No coreference resolution is performed outside sentences. Even if a given sentence comes from a document that allows us to resolve a coreference, as OIE is intended to be a task performed on isolated sentences, we only resolve the coreferences of entities included in sentences taken in isolation. Tuples using pronouns for which we can't identify the substitution element may seem meaningless, but coreference resolution must take place outside OIE, being a task in itself. In the example, we don't do coreference resolution for the pronoun \textit{He}, as no information about it is available in the sentence. However, we include a formulation replacing \textit{them} with \textit{tax reductions} in the annotation.

\begin{table}[htbp]
\centering
\begin{tabular}{|l|}
\hline
\textit{\begin{tabular}[c]{@{}l@{}}He did not go as far as he could have in\\tax reductions ; indeed he combined\\them with increases in indirect taxes .\end{tabular}} \\ \hline
\begin{tabular}[c]{@{}l@{}}\correct{(He \sep combined \textbf{them} with \sep}\\\textcolor{fecorrect}{increases in indirect taxes)}\\ \textcolor{fecorrect}{(He \sep combined \textbf{tax reductions} with \sep}\\\textcolor{fecorrect}{increases in indirect taxes)}\end{tabular}         \\ \hline
\end{tabular}
\caption*{Coreference resolution} 
\label{table:AnnoCoref}
\end{table}

\subsection{Inference}
Inference is necessary: facts directly implied by the sentence, even if not expressed verbatim, are relevant pieces of information. A nuance is necessary here with regard to potential implicit facts. These are not necessarily implied by the sentence and should therefore be omitted. In the first example, it is necessarily implied by the sentence that \textit{Paul Johanson} is \textit{Monsanto's Director of Science}. On the other hand, the fact that \textit{Monsanto's spray} is \textit{gentle on the female organ} is not necessarily true, what is true that this information is said by \textit{Paul Johanson}.

\begin{table}[htbp]
\centering
\begin{tabular}{|l|}
\hline
\textit{\begin{tabular}[c]{@{}l@{}}However , Paul Johanson , Monsanto ’s\\ director of plant sciences , said the\\company ’s chemical spray overcomes these\\problems and is gentle onthe female organ .\end{tabular}} \\ \hline
\begin{tabular}[c]{@{}l@{}}\correct{(Paul Johanson \sep is \sep}\\ \textcolor{fecorrect}{Monsanto's director of science)}\\ \\ \correct{(Paul Johanson \sep \textbf{says} \sep} \\\textcolor{fecorrect}{the company’s chemical spray is} \\\textcolor{fecorrect}{gentle on the female organ)}\\ \\ \wrong{(the company’s chemical spray \sep} \\\textcolor{fewrong}{is gentle on \sep} \\\textcolor{fewrong}{the female organ)}\end{tabular} \\ \hline
\end{tabular}
\caption*{Inference : First example}
\label{table:AnnoInf1}
\end{table}

It is then necessary to distinguish between light and heavy inference.  We define light inference as a form of inference that does not require logical reflection with respect to the sentence to deduce the fact, which is simply true as long as the sentence is also true. In the second example, the relation is implicit, but the annotated fact is obviously true. Heavy inference, on the other hand, requires some reflection or combination of logical operations to imply the fact. A case of heavy inference is that which requires external knowledge, as in the third example, where knowledge of human culture and the principle of heredity is necessary to make the inference. Another example of heavy inference is generalization, as in the fourth example, where a stronger fact is implied, a generalization of what is expressed in the sentence using a single example. A final example of heavy inference is that of lower or upper limits. As in the fifth example, we don't want to generalize lower or upper bounds to entities that are not directly expressed in the sentence. We therefore include in the reference facts that can be inferred using light inference, but not those resulting from heavy inference.

\begin{table}[htbp]
\centering
\begin{tabular}{|l|}
\hline
\textit{\begin{tabular}[c]{@{}l@{}}Jason Charles Beck, a Jewish Canadian\\  musician, was born in 1972.\end{tabular}} \\ \hline
\correct{(Jason Charles Beck \sep is \sep Jewish)}                                                                                      \\ \hline
\end{tabular}
\caption*{Inference : Second example}
\label{table:AnnoInf2}
\end{table}

\begin{table}[htbp]
\centering
\begin{tabular}{|l|}
\hline
\textit{\begin{tabular}[c]{@{}l@{}}Gonzales is the son of Ashkenazi Jews\\  who were forced to flee from Hungary\\  during World War II.\end{tabular}} \\ \hline
\wrong{(Gonzales \sep is \sep Jewish)  }                                                                                                                               \\ \hline
\end{tabular}
\caption*{Inference : Third example}
\label{table:AnnoInf3}
\end{table}

\begin{table}[htbp]
\centering
\begin{tabular}{|l|}
\hline
\textit{Gonzales is a McGill-trained virtuoso pianist.} \\ \hline
\wrong{(McGill \sep trains \sep pianists)}                              \\ \hline
\end{tabular}
\caption*{Inference : Fourth example}
\label{table:AnnoInf4}
\end{table}

\begin{table}[htbp]
\centering
\begin{tabular}{|l|}
\hline
\textit{\begin{tabular}[c]{@{}l@{}}The prefecture is part of the world's most\\  populous metropolitan area with upwards\\  of 37.8 million people and the world's\\  largest urban agglomeration economy.\end{tabular}} \\ \hline
\wrong{(the world \sep has \textbf{upwards} of \sep} \\\textcolor{fewrong}{37.8 million people)} \\ \hline
\end{tabular}
\caption*{Inference : Fifth example}
\label{table:AnnoInf5}
\end{table}

\subsection{Reformulation}

If a relation or argument is expressed in a complex way in the text, a simpler re-formulation of the same fact is added in the same cluster, even if the relation in the two formulations is not the same and the level of detail may be different. This is a compromise between the goal of OIE of collecting all the factual information expressed in the text and the importance of formulating these facts in simple language, which is relevant but not necessarily OIE's primary goal. The example shows a case where the reformulated relation is different but conveys the same meaning.

\begin{table}[htbp]
\centering
\begin{tabular}{|l|}
\hline
\textit{Sam managed to convince John}                                                             \\ \hline
\begin{tabular}[c]{@{}l@{}}\correct{(Sam \sep \textbf{managed to convince} \sep John)}\\ \textcolor{fecorrect}{(Sam \sep \textbf{convinced} \sep John)}\end{tabular} \\ \hline
\end{tabular}
\caption*{Reformulation}
\label{table:AnnoRefo}
\end{table}

\subsection{Active and Passive Voice}
Clusters are used to group together the active and passive formulations of an extraction in a single fact. If the active formulation is not present in the text, it should still be added in the same cluster if it is simpler than the original formulation present in the sentence. The example shows an originally passive tuple and it’s active formulation added in the same cluster.

\begin{table}[htbp]
\centering
\begin{tabular}{|l|}
\hline
\textit{The apple was eaten by Kyle}                                                             \\ \hline
\begin{tabular}[c]{@{}l@{}}\correct{(The apple \sep \textbf{was eaten by} \sep Kyle)}\\ \textcolor{fecorrect}{(Kyle \sep \textbf{ate} \sep the apple)}\end{tabular} \\ \hline
\end{tabular}
\caption*{Active and Passive Voice}
\label{table:AnnoAcPa}
\end{table}

\subsection{Attribution and Speculation}
Some information in the text is speculative or attributed to an entity, so this characteristic must be included in the relationship, in the way it is formulated in the sentence. This makes it possible to preserve this information without having to introduce a particular structure. This information must be included in the relation, as it is in no way related to the arguments. The example shows a case where the attribution is added in the relation of the tuple.

\begin{table}[htbp]
\centering
\resizebox{\columnwidth}{!}{%
\begin{tabular}{|l|}
\hline
\textit{The earth is flat, according to an Apple Valley man.}                                                             \\ \hline
\begin{tabular}[c]{@{}l@{}}\correct{(The earth \sep} \\\textcolor{fecorrect}{is \textbf{according to an Apple Valley man} \sep flat)}\\ \\ \wrong{(The earth \sep is \sep flat)}\end{tabular} \\ \hline
\end{tabular}%
}
\caption*{Attribution and Speculation}
\label{table:AnnoAttr}
\end{table}

\subsection{Correction}
Occasionally, some tuples may consist of words from the original sentence but contain grammatical errors. In this case, the tuple formed from the original words and the corrected tuple should be included in the same cluster. This ensures that neither the systems making the correction nor those using the original text are penalized. The example shows that \textit{newspaper} without an \textit{s} is a grammatical error, so both the original and the correction should be included.

\begin{table}[htbp]
\centering
\begin{tabular}{|l|}
\hline
\textit{\begin{tabular}[c]{@{}l@{}}He has written several newspaper and\\  magazine opinion pieces in The Guardian,\\  Vice, Billboard, and others.\end{tabular}} \\ \hline
\begin{tabular}[c]{@{}l@{}}\correct{(He \sep has written in \sep \textbf{newspaper})}\\ \textcolor{fecorrect}{(He \sep has written in \sep \textbf{newspapers})}\end{tabular}                                                        \\ \hline
\end{tabular}
\caption*{Correction}
\label{table:AnnoCorr}
\end{table}

% =====
\section{Matching Guidelines}
\label{sec:appendixmatchG}

This Appendix contains the matching guidelines for the open information extraction task and has been used in the development of the \(\bfl\) reference matching function. The various principles dictate which pairs of extractions made by systems and annotations should and should not match. The information is presented in the following format: Sentences are in the top cell and in the cell bellow, the different formulations of the same cluster (of the same fact) are in a paragraph and a line break separates them. Clusters in black represent annotations. Examples in \textcolor{fecorrect}{green}, preceded by a check mark are examples that match an annotation in the reference, while examples in \textcolor{fewrong}{red}, preceded by a cross mark do not.

\subsection{Exact match}
Two absolutely identical extractions should match.

\subsection{Relation specificity}
Extractions are allowed very little flexibility in the specificity of the relation: the relation is the vehicle of information, so it's important that it's almost as specific as the reference. That said, a different formulation that is just as specific should be accepted. In the example, \textit{was thrown} is not a relevant relation in the context of this extraction, as \textit{was} or \textit{was thrown out of} would have been (the word \textit{out} in argument 2 changes the meaning of the relation).

\begin{table}[htbp]
\centering
\begin{tabular}{|l|}
\hline
\textit{\begin{tabular}[c]{@{}l@{}}The Finns party was thrown out of\\ the government and  the new “Blue Reform”\\ group kept its cabinet seat.\end{tabular}} \\ \hline
\begin{tabular}[c]{@{}l@{}}(The Finns party \sep \textbf{was} \sep\\ thrown out of the government)\\ (The Finns party \sep \textbf{was thrown out of} \sep \\the government)\\ \\ \wrong{(The Finns party \sep \textbf{was thrown} \sep}\\ \textcolor{fewrong}{out of the government)}\end{tabular} \\ \hline
\end{tabular}
\caption*{Relation specificity}
\label{table:matchRelSpec}
\end{table}

\subsection{Errors}
Some extractions made by systems may present syntax or grammatical errors, when a word is misplaced or unnecessary. If this error changes the meaning of the relation or one of the arguments, the extraction should not be matched. If not, it should match the corresponding annotation. In the example, the word \textit{also} refers to the relation \textit{is}, and does not change the meaning of the relation, whereas the word \textit{and} changes the meaning of the extraction, making it nonsensical.

\begin{table}[htbp]
\centering
\begin{tabular}{|l|}
\hline
\textit{\begin{tabular}[c]{@{}l@{}}Known for his albums of classical piano\\  compositions, he is also a producer\\  and songwriter.\end{tabular}} \\ \hline
\begin{tabular}[c]{@{}l@{}}(He \sep is {[}also{]} \sep a songwriter)\\ \\ \wrong{(He \sep is \sep a songwriter \textbf{and})}\\ \\ \correct{(He \sep is \sep a songwriter \textbf{also})}\end{tabular}    \\ \hline
\end{tabular}
\caption*{Errors}
\label{table:matchErr}
\end{table}

\subsection{Word Choice}
Some words may be equivalent to those present in the annotations in certain contexts, even if we have chosen not to include them in the reference. If these words are used in the system extractions instead of those used in the reference, we still accept the system extraction. Some word choices may be wrong, but we still accept the extraction if the meaning remains. In the example, the determiner \textit{the} is used instead of \textit{a} in the extraction because it's the word found in the original sentence, but both are equally appropriate, so we accept the extraction.

\begin{table}[htbp]
\centering
\begin{tabular}{|l|}
\hline
\textit{\begin{tabular}[c]{@{}l@{}}He is the younger brother of the prolific\\  film composer Christophe Beck.\end{tabular}} \\ \hline
\begin{tabular}[c]{@{}l@{}}(He \sep is \sep \textbf{a} younger brother)\\ \\ \correct{(He \sep is \sep \textbf{the} younger brother)}\end{tabular}                      \\ \hline
\end{tabular}
\caption*{Word Choice}
\label{table:matchWorC}
\end{table}

\subsection{Level of Detail}
We want to match extractions which have a level of detail higher than the annotation but that convey the same information. By level of detail we mean that they combine information from two annotated clusters. On the other hand, if an extraction combines information from three or more annotated clusters, we consider it to be too noisy and not precise enough to be useful. The positive example is matched because it conveys the same information as the second annotated cluster, and only adds a single level of detail from the third cluster. The negative example is not matched because it combines information from all three annotated tuples into a long and imprecise second argument.

\begin{table}[htbp]
\centering
\resizebox{\columnwidth}{!}{%
\begin{tabular}{|l|}
\hline
\textit{\begin{tabular}[c]{@{}l@{}}Alex broadcasts a web series Music on a website.\end{tabular}} \\ \hline
\begin{tabular}[c]{@{}l@{}}(Alex \sep broadcasts \sep a web series)\\ \\ (Alex \sep broadcasts \sep Music)\\ \\ (Alex \sep broadcasts Music on \sep a website)\\ \\ \correct{(Alex \sep broadcasts \sep Music on a website)} \\ \\ \wrong{(Alex \sep broadcasts \sep }\\\textcolor{fewrong}{a web series Music on a website)}
\end{tabular} \\ \hline
\end{tabular}%
}
\caption*{Level of detail}
\label{table:matchLevDet}
\end{table}

\end{document}